\documentclass[lettersize,journal]{IEEEtran}
\usepackage{amsmath,amsfonts}
\usepackage{algorithmic}
\usepackage{algorithm}
\usepackage{array}
\usepackage[caption=false,font=normalsize,labelfont=sf,textfont=sf]{subfig}
\usepackage{textcomp}
\usepackage{stfloats}
\usepackage{url}
\usepackage{verbatim}
\usepackage{graphicx}
\usepackage{cite}
\usepackage{multirow}
\usepackage{xcolor}
\usepackage{booktabs,arydshln}

\usepackage{ulem}

\begin{document}

\title{Label Informed Contrastive Pretraining for Node Importance Estimation on Knowledge Graphs}

%\author{Tianyu Zhang, Chengbin Hou$^\dagger$, Rui Jiang, Xuegong Zhang, Chenghu Zhou, Ke Tang, \IEEEmembership{Fellow~IEEE}, Hairong Lv$^\dagger$
\author{Tianyu Zhang, Chengbin Hou$^\dagger$, Rui Jiang, Xuegong Zhang, Chenghu Zhou, Ke Tang, Hairong Lv$^\dagger$
        % <-this % stops a space
\thanks{$^\dagger$ Chengbin Hou and Hairong Lv are the corresponding authors.}
\thanks{E-mail: chengbin.hou10@foxmail.com; lvhairong@tsinghua.edu.cn}
\thanks{T. Zhang, R. Jiang, X. Zhang, and H. Lv are with the Ministry of Education Key Laboratory of Bioinformatics, Bioinformatics Division, Beijing National Research Center for Information Science and Technology, Department of Automation, Tsinghua University, Beijing, 100084, China}
\thanks{C. Hou is with the School of Computer Science and Engineering, Fuyao Institute of Technology, Fuzhou, 350109, China.}
\thanks{C. Zhou is with the State Key Laboratory of Resources and Environmental Information Systems, Institute of Geographic Sciences and Natural Resources Research, Chinese Academy of Sciences, Beijing 100101, China.}
\thanks{K. Tang is with the Department of Computer Science and Engineering, Southern University of Science and Technology, Shenzhen, 518055, China.}
\thanks{C. Hou and H. Lv are also with the Fuzhou Institute of Data Technology, Fuzhou, 350200, China.}
% <-this % stops a space
\thanks{Accepted by IEEE TNNLS}
\thanks{https://doi.org/10.1109/tnnls.2024.3363695}
\thanks{© 2024 IEEE. Personal use is permitted, but republication/redistribution requires IEEE permission. https://www.ieee.org/publications/rights/index.html}
%\thanks{Manuscript received xxxx xx, 2023; revised xxxx xx, 2023}
}

\markboth{Journal of \LaTeX\ Class Files,~Vol.~xx, No.~x, xxx~xxxx}%
{Shell \MakeLowercase{\textit{et al.}}: A Sample Article Using IEEEtran.cls for IEEE Journals}


\maketitle

\begin{abstract}
Node Importance Estimation (NIE) is a task of inferring importance scores of the nodes in a graph. Due to the availability of richer data and knowledge, recent research interests of NIE have been dedicating to knowledge graphs for predicting future or missing node importance scores. Existing state-of-the-art NIE methods train the model by available labels, and they consider every interested node equally before training. However, the nodes with higher importance often require or receive more attention in real-world scenarios, e.g., people may care more about the movies or webpages with higher importance. To this end, we introduce Label Informed ContrAstive Pretraining (LICAP) to the NIE problem for being better aware of the nodes with high importance scores. Specifically, LICAP is a novel type of contrastive learning framework that aims to fully utilize the continuous labels to generate contrastive samples for pretraining embeddings. Considering the NIE problem, LICAP adopts a novel sampling strategy called top nodes preferred hierarchical sampling to first group all interested nodes into a top bin and a non-top bin based on node importance scores, and then divide the nodes within top bin into several finer bins also based on the scores. The contrastive samples are generated from those bins, and are then used to pretrain node embeddings of knowledge graphs via a newly proposed Predicate-aware Graph Attention Networks (PreGAT), so as to better separate the top nodes from non-top nodes, and distinguish the top nodes within top bin by keeping the relative order among finer bins. Extensive experiments demonstrate that the LICAP pretrained embeddings can further boost the performance of existing NIE methods and achieve the new state-of-the-art performance regarding both regression and ranking metrics. The source code for reproducibility is available at \url{https://github.com/zhangtia16/LICAP}
%Extensive experiments on real-world knowledge graphs demonstrate the effectiveness of applying LICAP to existing NIE methods, as well as the benefits of proposed hierarchical sampling and PreGAT.
\end{abstract}


\section{Introduction}
\IEEEPARstart{N}{ode} Importance Estimation (NIE) aims to infer the significance or popularity of nodes in a graph, which has been applied to many applications such as web searching and resource allocation \cite{page1999pagerank,gomez2019centrality,park2019estimating,huang2021representation}. One of well-known methods to the NIE problem might be PageRank \cite{page1999pagerank}, which has been used in Google to rank web pages. PageRank employs random walks to simulate surfing the World Wide Web until convergence, so that the stationary probability of visiting a page can be assigned as its importance score. Besides, many random walk-based \cite{haveliwala2002topic,tong2008random} and centrality-based \cite{freeman2002centrality,borgatti2005centrality,borgatti2006centrality,gomez2019centrality} NIE methods have been proposed in the literature. It is worth noting that these early NIE methods are designed for the graph where nodes and edges belong to the same type respectively.

Due to the availability of richer structured data and knowledge \cite{bollacker2008freebase,suchanek2007yago,lehmann2015dbpedia,ji2021survey}, recent research interests of NIE have been dedicating to Knowledge Graphs (KG) for predicting future or missing importance scores of the interested nodes in a KG \cite{li2012har,park2019estimating,park2020multiimport,huang2021representation,huang2022hiven} in which nodes and edges belong to multiple types respectively, and the triplet (head node, relational edge, tail node) expresses real-world knowledge. \cite{li2012har} extends PageRank by incorporating multi-relational data to directly infer scores for both nodes and edges. More recently, the state-of-the-art performance has been achieved by the end-to-end training, which first builds a trainable model to include structure and/or knowledge in KG, and then optimizes the trainable model parameters using the labels of available node importance \cite{park2019estimating,park2020multiimport,huang2021representation}.

Despite the success of these trainable NIE methods on the real-world KG datasets, they treat every interested nodes equally before training. However, the potential real-world prior knowledge is that the nodes with higher importance often require or receive more attention, e.g., people may care more about the movies or webpages with higher importance. In other words, the correct importance estimation for higher importance nodes is usually more significant than that for lower importance nodes. As a result, this work tries to properly inject such prior knowledge into NIE methods.

\begin{figure}[htp]
\centering
\includegraphics[width=0.95\linewidth]{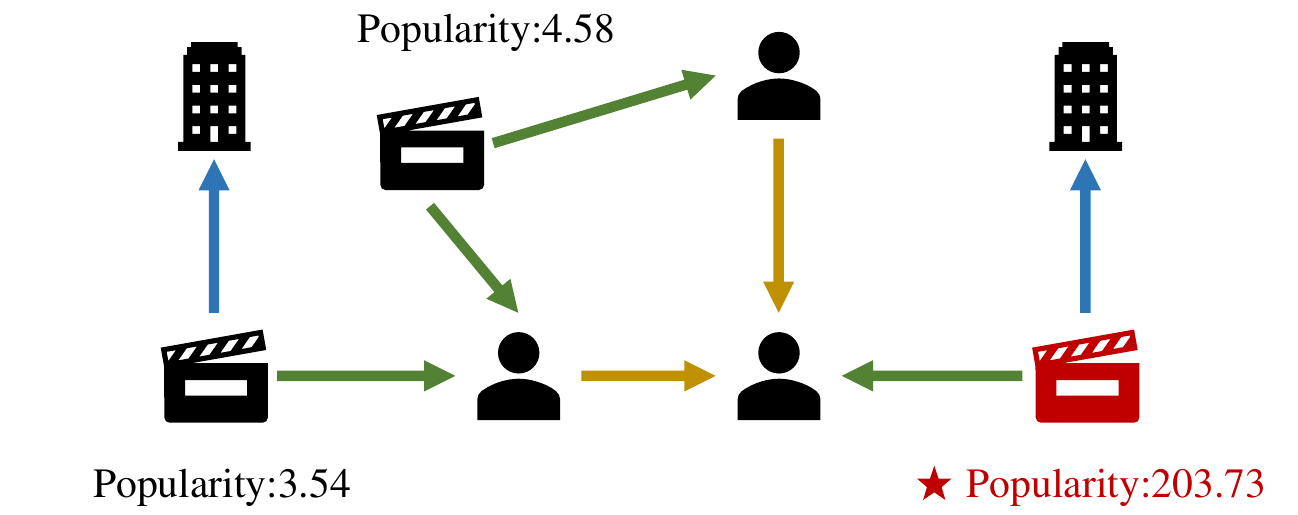}
\caption{The toy example of a movie KG. Edges with different colors indicate different edge types. Node importance of movies is defined by its popularity score. The movie (red) with higher importance often receives more attention.}
\label{fig:kg}
\end{figure}

Inspired by the recent advances of pretraining paradigm \cite{mikolov2013distributed,huh2016makes,kenton2019bert}, we introduce Label Informed ContrAstive Pretraining (LICAP) to inject the prior knowledge into NIE methods, so that the NIE methods can be better aware of the nodes with higher importance. As LICAP is introduced in pretraining stage, it can be easily collaborated with the existing NIE models that need the pretrained embeddings as the input features to the model.

Specifically, LICAP employs a novel type of contrastive learning framework that intends to make fully use of the continuous labels of samples. It is worth mentioning that contrastive learning has been mainly applied to discrete labels and classification problems, rather than continuous labels and regression problems. To this end, we transform the NIE regression problem into a classification-like problem by dividing the continuous labels of samples into bins and meanwhile maintaining the order among bins. 

To help NIE methods better realize the nodes with higher importance, LICAP adopts a novel sampling strategy called top nodes preferred hierarchical sampling to generate contrastive training samples. It first groups all interested nodes into a top bin and a non-top bin based on the continuous node importance scores. Second, the nodes within the top bin are further grouped into several finer bins also based on the scores. The contrastive samples for the two hierarchical grouping are generated respectively from the top bin against non-top bin and the finer bins within top bin. After that, a newly proposed Predicate-aware Graph Attention Networks (PreGAT) for knowledge graphs is applied to pretrain node embeddings, so as to better separate the top nodes from non-top nodes, and distinguish the top nodes within top bin by keeping the relative order among the finer bins. The pretrained embeddings, carrying the desirable prior knowledge, can be finally fed to the downstream NIE model.

The main contributions of this work are fivefold: (1) LICAP is introduced during pretraining stage for being better aware of higher important nodes, which can be easily collaborated with existing NIE methods. (2) To pretrain node embeddings via contrastive learning, a novel sampling strategy called top nodes preferred hierarchical sampling is proposed to utilize the continuous labels, so that the NIE regression problem is transformed into the classification-like problem in which the contrastive learning is applicable. (3) A new type of graph attention networks is developed to also include relational information, i.e., predicates, to better pretrain node embeddings on knowledge graphs. (4) Extensive experiments on three real-world knowledge graphs demonstrate that the LICAP pretrained embeddings can further boost the performance of existing NIE methods and achieve the new state-of-the-art performance regarding both regression and ranking metrics. (5) To benefit future research, we release the source code at \url{https://github.com/zhangtia16/LICAP}

%The organization of this paper is as follows. Section 2 ... Section 3 ...

\section{Related Work}
This work generally belongs to graph machine learning. The proposed method is introduced in pretraining stage, and the idea of constructive learning is used to pretrain embeddings. Thus, we survey the work for graph machine learning, and the graph related work for pretraining paradigm and contrastive learning. Finally, we focus on and review NIE methods.

\subsection{Graph Machine Learning}
Graph machine learning has attracted much attention from both academia and industry owing to the development of representation learning methods on graphs. DeepWalk \cite{perozzi2014deepwalk} and Node2Vec \cite{grover2016node2vec} are the two representative methods, which learn low dimensional node embeddings using random walks and simple neural networks, such that these embeddings can preserve some desirable properties from the input graph. More recently, GCN \cite{kipf2017semi}, GraphSAGE \cite{hamilton2017inductive}, GAT \cite{velivckovic2017gat} and many others are proposed by using either spectral or spatial graph convolution operation to deal with graph data in a more natural way. Overall, the recent graph machine learning methods are mainly based on neural networks to process graph data and generate embeddings for downstream tasks, and they are also extended to more complex graphs such as attributed graphs, heterogeneous graphs, and knowledge graphs \cite{schlichtkrull2018modeling,hou2020rosane,zhu2019relation,liu2021anomaly}. There is abundant literature on graph machine learning, and we refer to a comprehensive survey \cite{wu2020comprehensive} for further reading.

\subsection{Pretraining Paradigm}
The pretraining paradigm could trace back to pretrain deep learning models on a large labeled image dataset like ImageNet, and then fine-tune the pretrained model on the targeted dataset for computer vision tasks \cite{huh2016makes}. Later, the idea of self-supervising enables the researchers to pretrain deep learning models over massive unlabelled textual web data, and the pretrained model \cite{kenton2019bert} or word embeddings \cite{mikolov2013distributed} can boost the performance of various natural language processing tasks. The pretraining paradigm has also been applied in graph machine learning, e.g., Deepwalk \cite{perozzi2014deepwalk} and Node2Vec \cite{grover2016node2vec} can be seen as a kind of pretraining to obtain good node embeddings to facilitate downstream tasks on a graph; the variants of GIN \cite{xu2019GIN} and other models are employed to pretrain graph embeddings on a large-scale molecular graph dataset, and the pretrained model has been shown good generalizability to other molecular graph datasets \cite{xia2022survey}. In summary, we have witnessed the tremendous success of pretraining paradigm in many machine learning tasks. As a result, it is also worth attempting to inject the prior knowledge via pretraining in the NIE problem.

\subsection{Contrastive Learning}
% DGI (Deep Graph Infomax) MINE %(Mutual Information Neural Estimation) DIM %(Deep InfoMax)  Noise Contrastive Estimation(NCE)
Contrastive Learning (CL) is a recent surging research topic. Its main concept involves learning representations or embeddings by contrasting positive samples against negative ones. One classical CL loss is NCE \cite{oord2018representation}, which distinguishes one positive sample against one negative samples; while InfoNCE \cite{oord2018representation}, distinguishing one positive sample against several negative samples, has become the more popular CL loss. CL has achieved great success for representation learning in visual data. DIM \cite{hjelm2018learning} learns the representations of images by maximizing mutual information. Afterwards, many contrastive-based methods such as MoCo \cite{he2020momentum} and SimCLR \cite{chen2020simple} are proposed and achieve significant performance gains.
Encouraged by the success of CL in computer vision, CL has also been generalized to learn representations on graph-structured data. DGI \cite{velickovic2019deep} exploits DIM's mutual information maximization to learn the node-level representations, while InfoGraph \cite{sun2019infograph} uses the similar methodology as DGI to learn graph-level representations. Later, many other methods such as GraphCL \cite{you2020graph} and SimGCL \cite{yu2022graph} are proposed for graph contrastive learning. Previous work has demonstrated the benefits of CL in various machine learning problems; we thus adopt CL in the NIE problem to pretrain node embeddings by using InfoNCE loss to incorporate the prior knowledge of the preference for top nodes.

\subsection{Node Importance Estimation}
There have been many methods to address the NIE problem on different kinds of graphs ranging from homogeneous graphs to complicated heterogeneous graphs or knowledge graphs. The early NIE methods \cite{page1999pagerank,haveliwala2002topic,tong2008random,freeman2002centrality,borgatti2005centrality,borgatti2006centrality,gomez2019centrality} mainly consider homogeneous graphs, i.e., the graph with only a single node type and a single edge type. PageRank \cite{page1999pagerank} is a classical NIE method based on the random walk. By randomly traversing the graph and propagating the node importance, PageRank leverages the graph topology information for NIE. Personalized PageRank \cite{haveliwala2002topic}, a topic-sensitive version of PageRank, utilizes the specific topic information for NIE. Random Walk with Restart \cite{tong2008random} improves PageRank by introducing a restart mechanism. Apart from the random walk-based methods, many centrality-based methods have also been developed to identify important nodes. \cite{freeman2002centrality} formally define the concept of degree centrality and clarify the relation between node importance and centrality. Further discussion about the concept and measure of centrality can be found in \cite{borgatti2005centrality,borgatti2006centrality,gomez2019centrality}. However, all these methods are designed for homogeneous graphs, and thus can not capture richer information from heterogeneous graphs with multiple node types or edge types. 

Recently, to cope with more complicated graph data, researchers have developed some NIE methods for heterogeneous graphs or knowledge graphs \cite{li2012har,park2019estimating,park2020multiimport,huang2021representation,huang2022hiven,geng2022modeling}. 
HAR \cite{li2012har} implements random walks on multi-relational data, so that it can deal with homogeneous graphs. 
HIVEN \cite{huang2022hiven} is designed for heterogeneous information networks based on Graph Neural Networks (GNN) by tracing the local information of each node, i.e., the heterogeneity of the network.
DGNI \cite{geng2022modeling} is developed to predict future citation trends of newly published papers, which could be regarded as a NIE task on dynamic heterogeneous graphs.
GENI \cite{park2019estimating} is the first NIE method for knowledge graphs. To utilize the labelled data, GENI uses graph attention mechanism and neighbouring scores aggregation to infer node importance. 
MultiImport \cite{park2020multiimport} can handle multiple input signals and estimate  node importance on cross-domain knowledge graphs. 
RGTN \cite{huang2021representation} uses relational graph transformer to incorporate the node semantic information, i.e., the text descriptions for nodes, in the learning process. Notably, the GNN-based NIE methods such as GENI and RGTN have achieved the state-of-the-art performances on knowledge graphs.

Unlike previous NIE methods on knowledge graphs, this work introduces the potential real-world prior knowledge of the preference for top nodes (i.e., nodes with high importance scores). To inject the prior knowledge, a novel sampling strategy is proposed to incorporate contrastive learning for pretraining node embeddings using the available node importance scores. Also note that, the proposed LICAP is a kind of plug-in method and can be integrated to previous NIE methods, rather than a new specific NIE method.

\section{Preliminaries}
Unless stated otherwise, the symbols in bold $\mathbf{x}$ and $\mathbf{X}$ denote vectors and matrices respectively; the calligraphic uppercase symbols $\mathcal{X}$ denote sets; the operator on a set $|\mathcal{X}|$ counts its cardinal number.

\noindent\textbf{Definition 1.} 
(Knowledge Graph or KG): A knowledge graph $\mathcal{G=(V, E, P)}$ is a directed multi-relational heterogeneous graph with an edge mapping function $f_\mathcal{E}:\mathcal{E}\mapsto\mathcal{P}$, where nodes $v\in\mathcal{V}$, edges $e\in\mathcal{E}$, and edge types $p\in\mathcal{P}$ correspond to KG's entities, relations, and predicates respectively. Each relation in $\mathcal{E}$ belongs to one unique predicate in $\mathcal{P}$, and there can exist multiple relations between a pair of entities. The traditional homogeneous graph allows one edge type, whereas a KG often contains multiple edge types or predicates, i.e., $|\mathcal{P}|\gg1$, such that KG can hold comprehensive domain-specific knowledge. 

\noindent\textbf{Definition 2.}
(Node Importance): The importance of a node $v_i$ is given by its importance score $s_i \in \mathbb{R}^+$, i.e., a non-negative real number reflecting its significance or popularity of entities in a KG. For instance, the collected value of citation counts of a paper could serve as the node importance in an academic KG. Following \cite{huang2021representation,park2019estimating} to handle the skewed real-world importance values, the importance score $s_i$ used in our problem is obtained by log transformation of the collected real-world importance value. 

\noindent\textbf{Definition 3.}
(Node Importance Estimation or NIE): Given a knowledge graph $\mathcal{G=(V, E, P)}$ and a partial known node importance scores $\{s\}$ for the subset of interested nodes $\mathcal{V}_s \subseteq \mathcal{V}$, node importance estimation aims to learn a function $f:\mathcal{V}_s\mapsto\mathbb{R}^+$ that predicts the node importance score for every node in $\mathcal{V}_s$. To ensure that the importance scores for the nodes in $\mathcal{V}_s$ are comparable and meaningful, the nodes in $\mathcal{V}_s$ often belong to the same or similar node type.

\section{Methodology}
In this section, we present how to inject the prior knowledge into NIE methods for being better aware of higher important nodes. Section \ref{overview} illustrates the overview of the proposed framework and displays where to inject the prior knowledge. Section \ref{LICAP} elaborates the proposed LICAP in detail, which consists of several novel techniques to optimize the LICAP pretrained embeddings so that these embeddings can be better aware of higher important nodes than the original embeddings. Section \ref{GNN-NIE} shows how the LICAP pretrained embeddings can be incorporated with existing methods. Section \ref{algorithm} gives the implementation of LICAP and analyzes its theoretical complexity.

\subsection{Overview} \label{overview}
\begin{figure*}[htp]
\centering
\includegraphics[width=0.9\linewidth]{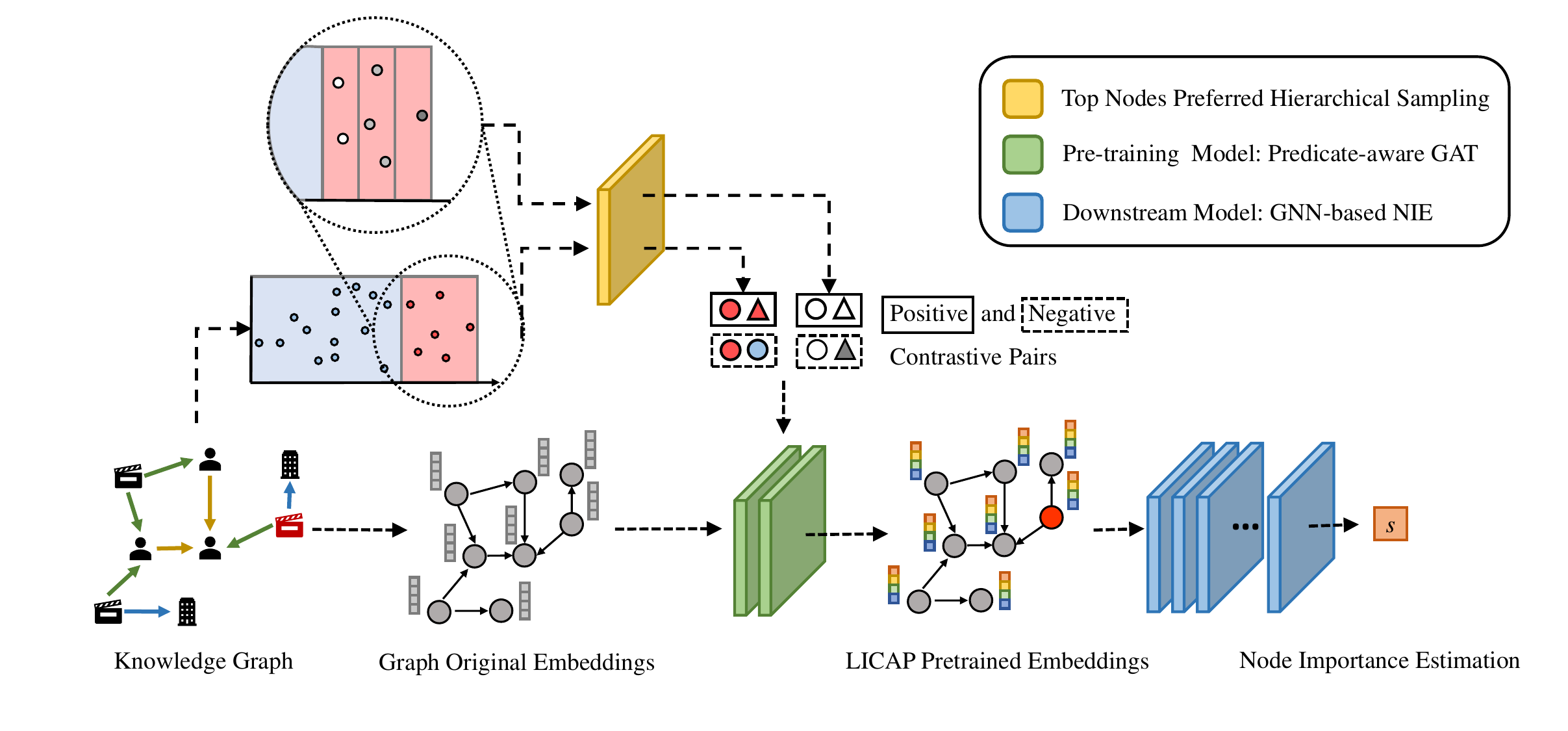}
\caption{An overview of the proposed framework.}
\label{fig:overview}
\end{figure*}

Fig.\ref{fig:overview} illustrates the overview of the proposed framework, which consists of the LICAP pretraining stage and the downstream GNN-based NIE training stage. 

Specifically, we introduce LICAP in pretraining stage to inject the prior knowledge of being better aware of higher important nodes. LICAP includes several key techniques or modules, which are presented in Section \ref{LICAP}. To be more specific, we first transform the NIE regression problem into a classification-like problem by the label informed grouping technique. Second, the top nodes preferred hierarchical sampling is proposed to generate contrastive samples based on the grouped bins as further illustrated in Fig.\ref{fig:l1}-\ref{fig:l2}. Third, these contrastive samples are fed to a newly proposed PreGAT, as shown in Fig.\ref{fig:pregat}, to pretrain node embeddings on knowledge graphs. After pretraining, we can feed the LICAP pretrained embeddings to the existing GNN-based NIE model, as discussed in Section \ref{GNN-NIE}, for downstream training and node importance estimation.

\subsection{Label Informed Contrastive Pretraining} \label{LICAP} 
The main novelties of this work lie in the proposed  Label Informed ContrAstive Pretraining (LICAP), which is elaborated in great details as follows.

\subsubsection{\textbf{Label Informed Grouping}}
To help NIE methods be better aware of the top nodes with higher importance scores during pretraining stage, it is straightforward to utilize the available node labels for pretraining node embeddings. In fact, the labels are continuous scores and it is indeed a regression problem. To some extent, the precise prediction of node importance scores is harder than the classification of node importance being roughly correct within a range, as the former needs to quantify to what degree they are different, while the latter only needs to qualitatively distinguish whether they are different. On the other hand, it might not be necessary to strongly correlate node embeddings with precise node scores in pretraining stage, since this would be done in downstream training stage. Consequently, we suggest to transform the NIE regression problem into a classification-like problem by using the idea of label
informed grouping technique.

Specifically, label informed grouping divides the continuous scores into a series of ordered bins, in which the contrastive pairs can be sampled based on these bins. The positive contrastive pairs are often taken from the same bin, whereas the negative contrastive pairs often come from the different bins. Because of also knowing the order of bins, the different degrees of negative contrastive pairs can be also informed. As such, the contrastive learning, which is not designed for regression problems, can be now adopted to use contrastive samples to pretrain embeddings.

\subsubsection{\textbf{Top Nodes Preferred Hierarchical Sampling and Contrastive Learning}}
To exploit contrastive learning for pretraining, we have to generate contrastive samples based on the bins given by label informed grouping over importance scores. To fully utilize the continuous importance scores, we develop a novel strategy called top nodes preferred hierarchical sampling. First, it groups all interested nodes into a top bin and a non-top bin based on the scores; accordingly, the first level of contrastive samples can be generated from the top bin and non-top bin. Second, all the nodes in the top bin are further grouped into several finer bins also based on the scores; accordingly, the second level of contrastive samples can be generated from these finer bins within the top bin. Next, we interpret the sampling of two levels of contrastive samples along with corresponding contrastive losses.

\begin{figure}[htbp]
\centering
\includegraphics[width=1.00\linewidth]{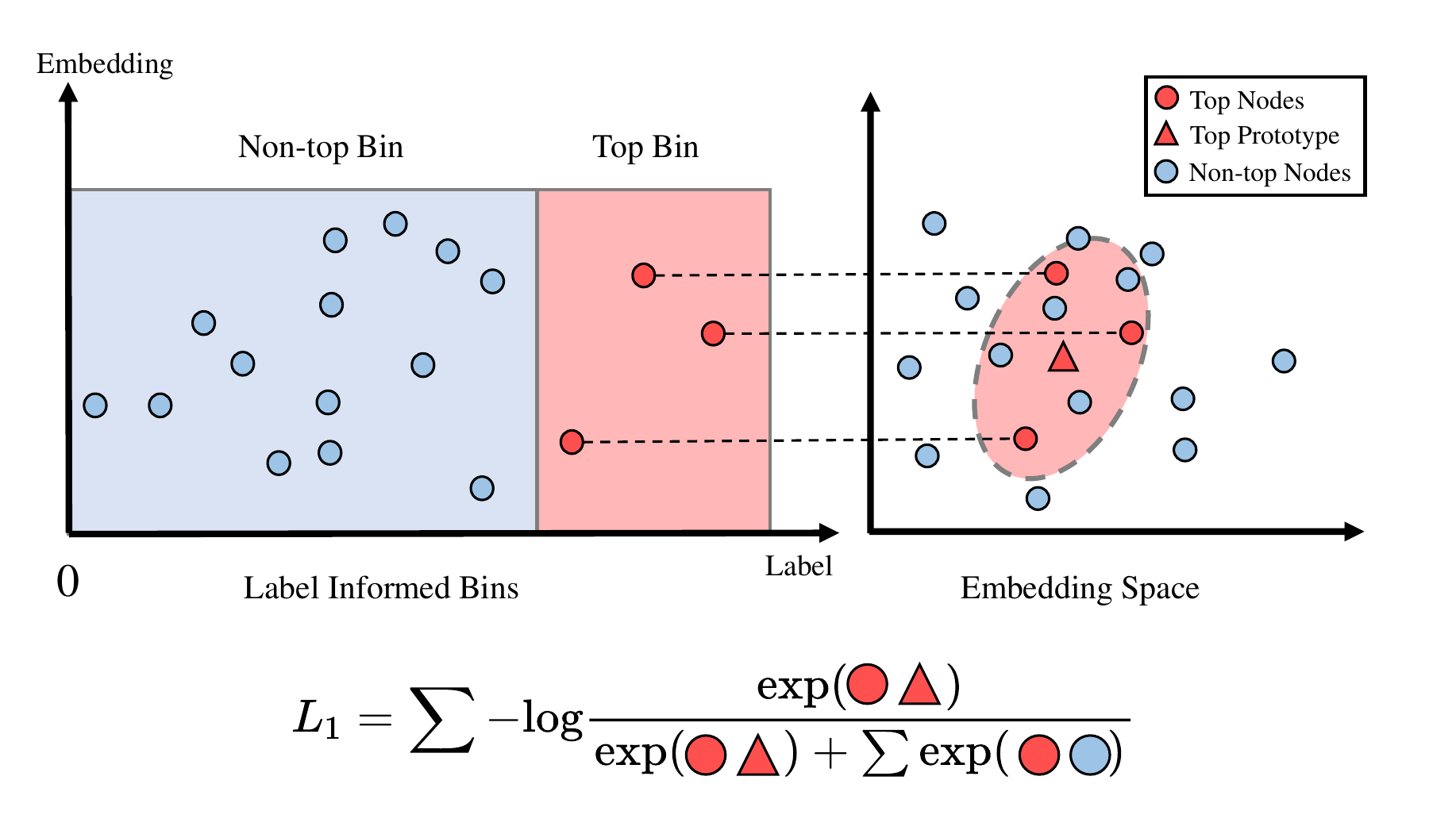}
\caption{The contrastive samples and loss for the top bin and non-top bin to better separate top nodes from non-top nodes.}
\label{fig:l1}
\end{figure}
As shown in Fig.\ref{fig:l1}, we first split all samples into two parts along the label dimension. The red part is the top bin accommodating top nodes, while the blue one is the non-top bin accommodating non-top nodes. The ratio for splitting is called important ratio $\gamma$. To better capture the common characteristics of top nodes, we suggest to contrast each node in the top bin positively to the top prototype (i.e., the triangle in Fig.\ref{fig:l1}). The top prototype is obtained by element-wisely averaging the embeddings of all top nodes. Regarding negative contrast, we treat the non-top nodes, which are randomly sampled from the non-top bin, as the negative pairs to each top node. The contrastive loss over each node in the top bin can be written as:
%\begin{equation}
%  L_1=\sum_{i\in \{top\}}-\text{log}\frac{\text{exp}(\mathbf{h}_i\cdot \mathbf{c}^{\text{bin}}/\tau)}{\sum_{j\notin \{top\}}\text{exp}(\mathbf{h}_i\cdot \mathbf{h}_j/\tau)} 
%  \label{eq:loss1-old}
%\end{equation}
\begin{equation}
  L_1=\sum_{i\in \text{bin}_\text{top}}-\text{log}\frac{\text{exp}(\mathbf{h}_i\cdot \mathbf{c}_{\text{top}}/\tau)}{\text{exp}(\mathbf{h}_i\cdot \mathbf{c}_{\text{top}}/\tau)+\sum_{j \notin \text{bin}_\text{top}}\text{exp}(\mathbf{h}_i\cdot \mathbf{h}_j/\tau)} 
  \label{eq:loss1}
\end{equation}
where $\mathbf{h}_i$ is the output embedding of pretraining model as introduced later in Section \ref{sec:PreGAT}; $\text{bin}_\text{top}$ contains all top nodes in the top bin; $\mathbf{c}_{\text{top}}$ is the top prototype for $\text{bin}_\text{top}$; $\tau$ is the temperature parameter of InfoNCE \cite{oord2018representation}.

\begin{figure}[hbtp]
\centering
\includegraphics[width=1.00\linewidth]{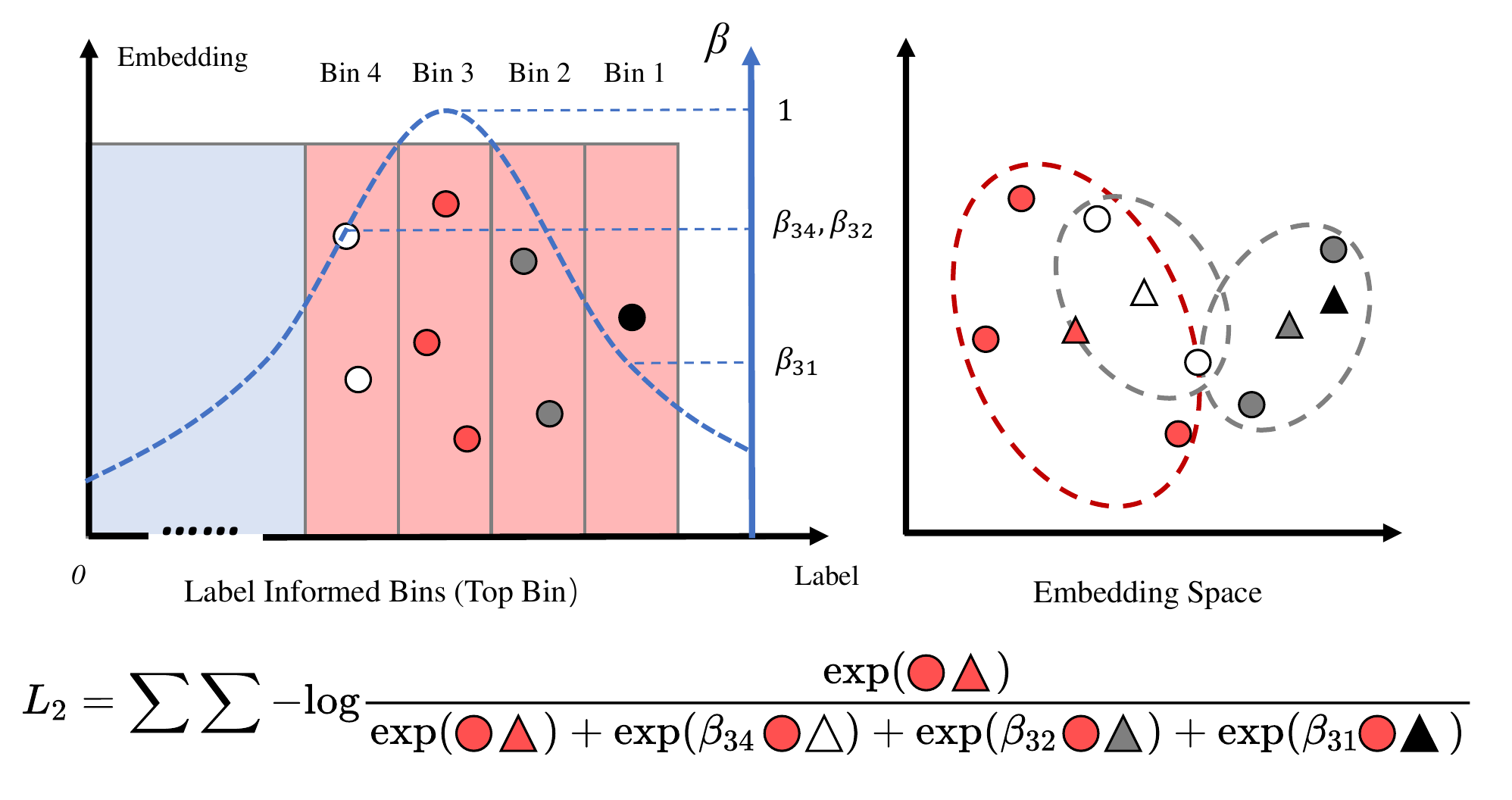}
\caption{The contrastive samples and loss for the finer bins within top bin to better distinguish top nodes by keeping relative orders among the finer bins.}
\label{fig:l2}
\end{figure}
To better distinguish the top nodes within the top bin, the second level of contrastive samples is generated based on the finer bins within the top bin. As shown in Fig.\ref{fig:l2}, the finer bins are also constructed according to the continuous label dimension. We then contrast each top node to the corresponding positive and negative samples. For each top node in a finer bin, its positive contrastive sample is defined as the finer bin's prototype (e.g., the red triangle for the finer bin holding red nodes in Fig.\ref{fig:l2}), which is likewise obtained by element-wisely averaging the embeddings of the nodes in this finer bin. Regarding the negative contrastive samples that are also prototypes for remaining finer bins respectively, we design a re-weighting negative sampling mechanism to keep the relative order among these finer bins. The re-weighting negative sampling mechanism is based on the intuition that the samples from farther apart bins are more dissimilar (or less proximate). As a result, we introduce a proximate coefficient $\beta$ to re-weight the negative contrastive pairs, such that the larger the proximate coefficient is, the greater the impact of the corresponding negative pairs term appears in the denominator. The contrastive loss over each node in (the finer bins within) the top bin is as follow:
%\begin{equation}
%  L_2=\sum_{i\in \{top\}}-\text{log}\frac{\text{exp}(\mathbf{h}_i\cdot \mathbf{c}^{\text{bin}}_i/\tau)}{\sum_{j\in \{prototypes\}}\text{exp}(\beta _{ij}\cdot \mathbf{h}_i\cdot \mathbf{c}^{\text{bin}}_j/\tau)}  
%  \label{eq:loss2-old}
%\end{equation}
\begin{equation}
  L_2=\sum_{m=1}^{B}\sum_{i\in \text{bin}_m}-\text{log}\frac{\text{exp}(\mathbf{h}_i\cdot \mathbf{c}_m/\tau)}{\sum_{n=1}^{B}\text{exp}(\beta _{mn}\cdot \mathbf{h}_i\cdot \mathbf{c}_n/\tau)}  
  \label{eq:loss2}
\end{equation}
where $\text{bin}_m$ contains all nodes in the $m$-th finer bin; $B$ is the total number of finer bins; $\bigcup\nolimits_{m=1}^B \text{bin}_m = \text{bin}_\text{top}$ and $\text{bin}_m \cap \text{bin}_n = \emptyset$, $\forall~~m \neq n$; $\mathbf{c}_m$ is the prototype of $\text{bin}_m$ to which $\mathbf{h}_i$ belongs; $\tau$ is the temperature parameter; $\beta _{mn}$ is the proximity coefficient between $m$-th and $n$-th finer bins. Notably, $\beta$ is subject to a binomial distribution:
\begin{equation}
  \beta \sim \mathbb{B}~(N, p)
  \label{eq:beta}
\end{equation}

To be more specific, the probability function of the binomial distribution is $P(X=k)=C^{k}_{N}~p^k(1-p)^{N-k}$. To gain a symmetric distribution, we take $p=0.5$. We then choose $N=2~\text{max}(m,B-m)$ so that $N$ is even and such binomial distribution only has one unique maximum. After substituting $p$ and $N$ into the the probability function, we derive $\beta_{mn} = C^{n-m+N/2}_{N} (0.5)^N$. Next, we normalized $\beta_{mn}$ so that its maximum is 1 when $m=n$. The formulation of $\beta_{mn}$ for re-weighting contrastive samples in the denominator of Equation \ref{eq:loss2} finally becomes: 
\begin{equation}
  \beta_{mn} = \frac{C^{n-m+N/2}_{N} }{C^{N/2}_{N}}
  \label{eq:beta-version3}
\end{equation}

It is worth noting the first level of contrastive samples and $L_1$ loss aim to better separate the top nodes from non-top nodes, while the second level of contrastive samples and $L_2$ loss try to better distinguish the top nodes within the top bin by keeping the relative order among the finer bins. As a result, it becomes possible to be better aware of the nodes with higher importance scores. The two losses are jointly optimized (Section \ref{sec:opt}) to train embeddings $\mathbf{h}_i$ by using PreGAT (Section \ref{sec:PreGAT}) as the pretraining model.

\subsubsection{\textbf{Predicate-aware GAT}} \label{sec:PreGAT}
To incorporate predicates that are also valuable information for knowledge graphs, we develop Predicate-aware Graph Attention Networks (PreGAT) as the pretraining model for LICAP to update embeddings $\mathbf{h}_i$.

\begin{figure}[hbtp]
\centering
\includegraphics[width=0.88\linewidth]{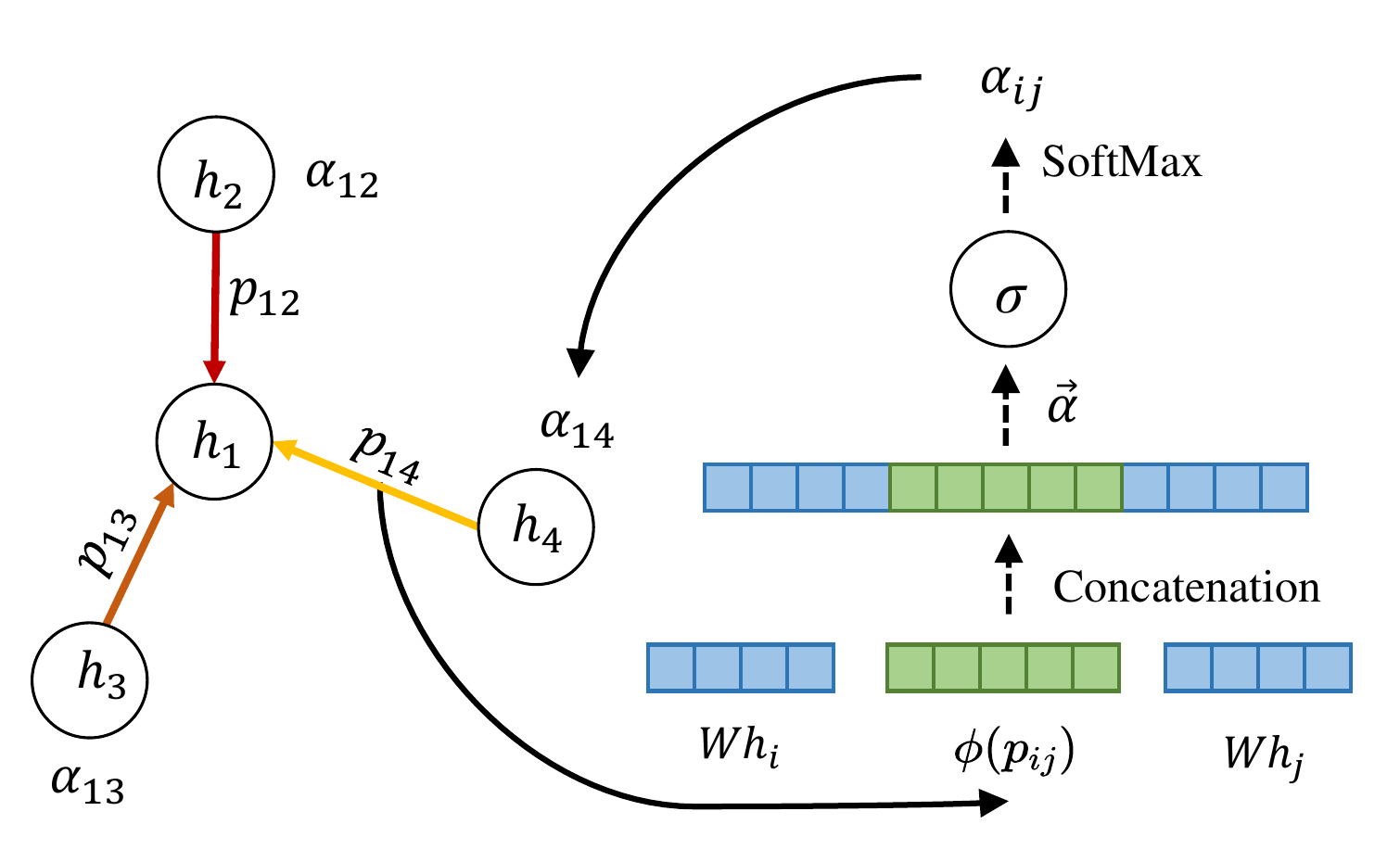}
\caption{Predicate-aware GAT (PreGAT) to inherently include predicates (i.e., edge types containing relational information) in knowledge graphs.}
\label{fig:pregat}
\end{figure}

As illustrated in Fig.\ref{fig:pregat}, PreGAT is modified from the classical GAT \cite{velivckovic2017gat}. Comparing to GAT, PreGAT considers the predicate in the concatenation step when computing attention coefficients. Let $p_{ij}$ denote the predicate (i.e., edge type) index between node $i$ and $j$. A trainable mapping function $\phi(\cdot)$ is used to transform the predicate index $p_{ij}$ into its corresponding embedding. The embeddings of predicates are initialized from the normal distribution $\mathcal{N}(0,1)$ and learned by back propagation. It is noteworthy that the predicates are considered as categorical in this work. There may exist knowledge graphs whose predicates hold ordinal or hierarchical nature, which is left as future work. The whole concatenation vector is fed to a single layer neural network with a non-linearity mapping $\sigma$, and then normalized via softmax. Formally, PreGAT calculates the attention coefficients $\alpha_{ij}$ of node $i$ on $j$ as follows:
\begin{equation}
  \alpha_{ij}=\frac{\text{exp}(\sigma(\mathbf{a} \cdot [\mathbf{W}\mathbf{h}_i\Vert\phi(p_{ij})\Vert\mathbf{W}\mathbf{h}_j]))}{\sum_{k\in \mathcal{N}_{i}}^{}\text{exp}(\sigma(\mathbf{a} \cdot [\mathbf{W}\mathbf{h}_i\Vert\phi(p_{ik})\Vert\mathbf{W}\mathbf{h}_k]))}
  \label{eq:alpha_rel}
\end{equation}
where the non-linearity mapping $\sigma$ takes LeakyReLU as used in \cite{velivckovic2017gat}; $\mathbf{a}$ is the parameterized weight vector; $\mathbf{W}$ is the transformation weight matrix; $\Vert$ represents concatenation.

%\begin{equation}
%  \alpha_{ij}=\frac{\text{exp}(\sigma_\alpha(\mathbf{\alpha}^{T}[\mathbf{W}\mathbf{h}_i\Vert\phi(p_{ij})\Vert\mathbf{W}\mathbf{h}_j]))}{\sum_{k\in \mathcal{N}_{i}}^{}\text{exp}(\sigma_\alpha(\mathbf{\alpha}^{T}[\mathbf{W}\mathbf{h}_i\Vert\phi(p_{ij})\Vert\mathbf{W}\mathbf{h}_k]))}
%  \label{eq:alpha_rel}
%\end{equation}

Similar to the original GAT, PreGAT learns the node embedding by aggregating the embeddings of its neighbours with corresponding attention coefficients, i.e.,
\begin{equation}
  \mathbf{h}_i=\sigma(\sum_{k\in \mathcal{N}_{i}}^{}\alpha_{ik}\mathbf{W}\mathbf{h}_k)
  \label{eq:hi}
\end{equation}
where $\mathcal{N}_{i}$ denotes the neighbors of node $i$, $\alpha_{ik}$ is the attention coefficient of node $i$ on $k$; $\mathbf{W}$ is the shared transformation weigh matrix; the non-linearity $\sigma$ also takes LeakyReLU.

Following \cite{velivckovic2017gat}, a multi-head attention mechanism is also deployed in our PreGAT. The output embeddings from each head are concatenated together to gain the final output:
\begin{equation}
  \mathbf{h}_i= \mathop{\Vert}_{h=1}^H \sigma(\sum_{k\in \mathcal{N}_{i}}^{}\alpha_{ik}^h\mathbf{W}^h\mathbf{h}_k)
  \label{eq:hihead}
\end{equation}
where $\Vert$ indicates a concatenation operator; $\alpha_{ik}^h$ is the attention coefficient calculated by the $h$-th head; the final output $\mathbf{h}_i$ consists of features learned from $H$ heads.

\subsubsection{\textbf{Optimization}} \label{sec:opt}
To be better aware of the nodes with higher importance and exploiting the predicates of knowledge graphs during pretraining stage, we update the node embeddings $\mathbf{h}_i$ using PreGAT and optimize $\mathbf{h}_i$ by jointly minimizing the two levels of contrastive losses $L_1$ and $L_2$. The overall objective for the optimization of LICAP pretraining is as follows:
%\mathcal{L}=L_1+L_2
\begin{equation}
  L=\eta_1~L_1+\eta_2~L_2
  \label{eq:totalloss}
\end{equation}
where $\eta_1$ and $\eta_2$ are two hyper-parameters to balance the two losses respectively. We adopt Adam optimizer and employ the early stopping strategy to alleviate overfitting. After pretraining, we obtain the pretrained node embeddings $\mathbf{h}_i^{\text{LICAP}}$, which can be then used in downstream tasks.

\subsection{Downstream GNN-based NIE}\label{GNN-NIE}
To make use of the existing NIE methods and incorporating with LICAP pretrained embeddings, we reformulate the existing GNN-based NIE methods. Given the pretrained embeddings $\mathbf{h}_i^{\text{LICAP}}$ for node $i$, the downstream GNN-based NIE model following the neighborhood aggregation scheme can be described as:
\begin{equation}
\mathbf{m}_{ij}^{(t+1)} = \text{aggregate}\left( \{ \mathbf{h}_i^{(t)}, \mathbf{h}_j^{(t)}, \mathbf{e}_{ij}^{(t)}: j\in \mathcal{N}_i \} \right)
  \label{eq:aggregate}
\end{equation}
\begin{equation}
\mathbf{h}_i^{(t+1)} = \text{update}\left( \mathbf{h}_i^{(t)}, \mathbf{m}_{ij}^{(t+1)}\right)
  \label{eq:update}
\end{equation}
where the initial node feature $\mathbf{h}^0_i$ or $\mathbf{h}^0_j$ takes the LICAP pretrained node embeddings for node $i$ or $j$; the edge feature $\mathbf{e}_{ij}$ denotes the optional edge embeddings; the operator $\text{"aggregate"}$ is the differentiable and permutation invariant function (e.g., sum, mean, max) to combine incoming messages $\{\cdot\}$ at step $t$ from the neighbors of $i$ given by $\mathcal{N}_i$; the operator $\text{"update"}$ is a differentiable function (e.g., neural networks) to update embeddings for node $i$ at step $t+1$. This neighborhood aggregation scheme (a.k.a. message passing scheme) allows GNN to iteratively update the node features based on the neighboring features.

Regarding the NIE problem on knowledge graphs, the state-of-the-art performance has been achieved by GENI \cite{park2019estimating} and RGTN \cite{huang2021representation}, which are both based on the above GNN neighborhood aggregation scheme. Concretely, GENI calculates the initial scores via $s_i^0 = f_{\text{score-GENI}}(\mathbf{h}_i^{\text{LICAP}})$ and then follows the above scheme to aggregate and update scores by $s_i = f_{\text{GNN-GENI}}(s_i^0,s_j^0,e_{ij}^0:j\in\mathcal{N}_i)$, while RGTN follows the above scheme to aggregate and update embeddings via $\mathbf{h}_i = f_{\text{GNN-RGTN}}(\mathbf{h}_i^{\text{LICAP}},\mathbf{h}_j^{\text{LICAP}},\mathbf{e}_{ij}^{\text{LICAP}}:j\in\mathcal{N}_i)$ and then calculates the scores by $s_i = f_{\text{score-RGTN}}(\mathbf{h}_i)$. Notably, there could be multiple layers of $f_{\text{GNN-GENI}}$ and $f_{\text{GNN-RGTN}}$ respectively, and both $f_{\text{score-GENI}}$ and $f_{\text{GNN-RGTN}}$ can be modeled as simple neural networks to project embeddings into scores. After reformulating GENI and RGTN with the general GNN neighborhood aggregating scheme, we can incorporate the LICAP pretrained embeddings into these GNN-based NIE models for node importance estimation.

\subsection{Algorithm and Complexity Analysis} \label{algorithm}

\renewcommand{\algorithmicrequire}{\textbf{Input:}}
\renewcommand{\algorithmicensure}{\textbf{Output:}}
\renewcommand{\algorithmiccomment}[1]{\hfill $\triangleright$ #1}

\begin{algorithm}[tb]
\caption{Label Informed Contrastive Pretraining}\label{alg:algorithm1}
\begin{algorithmic}[1]

    \REQUIRE Knowledge graph $\mathcal{G=(V, E, P)}$; Initial node embedding $\mathbf{e}\in\mathbf{R}^{|\mathcal{V}|\times F}$; Important ratio $\gamma$; Number of finer bins $B$; Balancing parameters for losses $\eta_1$ and $\eta_2$; Temperature parameter $\tau$; PreGAT $f_\Theta$($\cdot$)
    \ENSURE LICAP pretrained embedding $\mathbf{h}^{\text{LICAP}}\in\mathbf{R}^{|\mathcal{V}|\times F'}$

    \STATE initialize $\Theta$
    \STATE group nodes into $\text{bin}_\text{top}$ by $\gamma$ and labels' order
    \STATE group nodes in $\text{bin}_\text{top}$ into $\{\text{bin}_\text{1}\dots\text{bin}_\text{B}\}$  by labels
    %\COMMENT{Label informed grouping}
    \STATE obtain proximate coefficient $\{\beta_{11}\dots\beta_{BB}\}$ \COMMENT{Eq.~(\ref{eq:beta-version3})} 
    
    \FOR  {\emph{number of epochs}}
    \STATE $\mathbf{h} = f_\Theta(\mathcal{G},{\mathbf{e}})$ 
    \COMMENT{Eq.~(\ref{eq:alpha_rel})-(\ref{eq:hihead})} 

    \STATE compute $\mathbf{c}_{\text{top}}$ by aggregating $\mathbf{h}$ of nodes in $\text{bin}_\text{top}$
    \STATE compute $\{\mathbf{c}_1\dots\mathbf{c}_B\}$ by aggregating $\mathbf{h}$ of nodes in $\{\text{bin}_\text{1}\dots\text{bin}_\text{B}\}$ respectively   \COMMENT{parallel} 
    
    \STATE initialize $L_1=0$ and $L_2=0$

    \STATE obtain $\{\mathbf{h}_j\}$ by randomly sampling node $j\notin\text{bin}_\text{top}$
    
    \FOR {node $i$ in $\text{bin}_\text{top}$}
 
    % \STATE $L_1 \leftarrow L_1 -\text{log}\frac{\text{exp}(\mathbf{h}_i\cdot \mathbf{c}_{\text{top}}/\tau)}{\text{exp}(\mathbf{h}_i\cdot \mathbf{c}_{\text{top}}/\tau)+\sum_{j \notin \text{bin}_\text{top}}\text{exp}(\mathbf{h}_i\cdot \mathbf{h}_j/\tau)}$
    \STATE $l_{1,i} = -\text{log}\frac{\text{exp}(\mathbf{h}_i\cdot \mathbf{c}_{\text{top}}/\tau)}{\text{exp}(\mathbf{h}_i\cdot \mathbf{c}_{\text{top}}/\tau)+\sum_{j \notin \text{bin}_\text{top}}\text{exp}(\mathbf{h}_i\cdot \mathbf{h}_j/\tau)}$
    \STATE $L_1 \leftarrow L_1 + l_{1,i}$
    \COMMENT{Eq.~(\ref{eq:loss1})} 
    
    \ENDFOR
    %\STATE $L_1 = \sum  l_{1,i}$

    \FOR {$m$ = 1 to $B$} 
        \FOR {node $i$ in $\text{bin}_m$}
        \STATE $l_{2,i} = -\text{log}\frac{\text{exp}(\mathbf{h}_i\cdot \mathbf{c}_m/\tau)}{\sum_{n=1}^{B}\text{exp}(\beta_{mn}\cdot \mathbf{h}_i\cdot \mathbf{c}_n/\tau)}$
        \STATE $L_2 \leftarrow L_2 + l_{2,i}$
        \COMMENT{Eq.~(\ref{eq:loss2})} 
        \ENDFOR
    \ENDFOR

    \STATE $L=\eta_1~L_1+\eta_2~L_2$
    \COMMENT{Eq.~(\ref{eq:totalloss})}
    \STATE update $\Theta$ by minimizing $L$ via gradient descent

    \ENDFOR
    
    \STATE $\mathbf{h}^{\text{LICAP}} = f_\Theta(\mathcal{G},{\mathbf{e}})$
    \COMMENT{Eq.~(\ref{eq:alpha_rel})-(\ref{eq:hihead})} 
    
    \RETURN $\mathbf{h}^{\text{LICAP}}$

\end{algorithmic}
\end{algorithm}

As shown in Section \ref{LICAP} and Algorithm \ref{alg:algorithm1}, LICAP consists of several stages. 
During the stage of label informed grouping, ranking nodes in line 2 requires $\mathcal{O}(|\mathcal{V}|\text{log}|\mathcal{V}|)$ and grouping nodes into finer bins in line 3 requires $\mathcal{O}(\gamma|\mathcal{V}|)$. When it comes to proximate coefficient in line 4, the complexity is $\mathcal{O}(B^2)$.
During the stage of PreGAT's encoding, as explained in \cite{velivckovic2017gat}, the time complexity of GAT is $\mathcal{O}(FF'|\mathcal{V}|+F'|\mathcal{E}|)$ where $F$ and $F'$ are the dimension of input and output features respectively. For the multi-head mechanism, though multiplying the spatial complexity by a factor of $H$, GAT's time complexity is not affected since the computations of multiple heads can be parallelized. Similarly, the time complexity of proposed PreGAT is $\mathcal{O}(FF'|\mathcal{V}|+(F'+P')|\mathcal{E}|)$, corresponding to line 6 and 24. $P'$ is the dimension of predicate embedding $\phi(p)$. It is noted that the additional cost of PreGAT over GAT is relatively trivial, since $P'\le F'$ in most cases. 
During the stage of hierarchical sampling and contrastive learning, the main cost is at calculating two losses. 
Regarding $L_1$, the complexity of negative sampling in line 10 is $\mathcal{O}(k_{neg}(1-\gamma)|\mathcal{V}|)$, where negative sampling ratio $k_{neg}<1$ and important ratio $\gamma<1$. The complexity is $\mathcal{O}(\gamma|\mathcal{V}|)$ for line 11. Thus, the total complexity of calculating $L_1$ is $\mathcal{O}((k_{neg}(1-\gamma)+\gamma)|\mathcal{V}|)$, which corresponds to line 10-14.
Regarding $L_2$, the complexity is $\mathcal{O}(\gamma|\mathcal{V}|)$ for line 15-16, which yields the total complexity of calculating $L_2$ to be $\mathcal{O}(\gamma|\mathcal{V}|)$ corresponding to line 15-20.
Overall, the complexity of LICAP is $\mathcal{O}(|\mathcal{V}|\text{log}|\mathcal{V}|+\gamma|\mathcal{V}|+\mathcal{O}(B^2)) + t (\mathcal{O}(FF'|\mathcal{V}|+(F'+P')|\mathcal{E}|+(k_{neg}(1-\gamma) + 2\gamma)|\mathcal{V}|))$ where $t$ is the number of epochs. To sum up, LICAP is proportional to  $|\mathcal{V}|\text{log}|\mathcal{V}|$ and linearly proportional to $|\mathcal{E}|$.

%quadratic\footnote{It might be worth noting that the quadratic term of $\mathcal{O}(k_{neg}\gamma(1-\gamma)|\mathcal{V}|^2)$ is multiplied by three factors smaller than 1.}

\section{Experimental Settings}
This section presents datasets, baseline methods, evaluation metrics, and implementation details. Comparing to the previous work of NIE in knowledge graphs, a new dataset GA16K and additional regression metrics are suggested for benchmark.

\subsection{Datasets}
To evaluate the proposed methodology, we conduct experiments over three real-world knowledge graphs. The statistics of them are shown in TABLE \ref{tab_statistics}. 
The importance scores, acting as the labels, are obtained by log transformation of the collected real-world importance value such as pageviews for FB15K and citation counts for GA16K, so as to handle the skewed real-world importance values \cite{huang2021representation,park2019estimating}.
\begin{itemize}
\item \textbf{FB15K} is a subset of Freebase\footnote{\url{http://www.freebase.be/}}. It has around 15,000 entities with 1,345 different predicates in the form of knowledge base relation triples. The pageviews of Wikipedia pages are taken as the node importance.

\item \textbf{TMDB5K} is a movie knowledge graph generated from TMDB 5000 dataset\footnote{\url{https://www.kaggle.com/tmdb/tmdb-movie-metadata}}. It includes around 5,000 movie entities as well as other entities such as actors, directors, and countries. The popularity scores of movies are used as the node importance. 

\item \textbf{GA16K} is a Geoscience academic knowledge graph from GAKG\footnote{\url{https://gakg.acemap.info/}}. GAKG \cite{deng2021gakg} is a large-scale multimodal academic knowledge graph based on geoscience-related papers. GA16K contains about 16,000 entities such as papers, authors, and affiliations. The citation counts of papers are defined as the node importance.

%\item \textbf{IMDB} is another movie knowledge graph built from IMDB dataset\footnote{\url{https://www.imdb.com/interfaces/}}. It consists of about 1.1 million nodes and 9.7 million edges. The nodes or entities contain movies, genres, directors, casts, crews, etc. The votes of movies are taken as the node importance. %It is noteworthy that IMDB is the largest knowledge graph among the three datasets.
\end{itemize}

\begin{table}[htbp]
  \centering
  \caption{Statistics of real-world knowledge graphs.}
    \scalebox{0.93}{
    \begin{tabular}{lrrrr}
    \toprule
    Datasets & Nodes & Edges & Predicates & Nodes with Importance \\
    \midrule
    FB15K    & 14,951    & 592,213    & 1,345   & 14,105 (94.3\%)   \\
    GA16K    & 16,363    & 151,662    & 10      & 8,966 (54.8\%)    \\
    TMDB5K   & 114,805   & 761,648    & 34      & 4,803 (~4.2\%)    \\
    %IMDB     & 1,124,995 & 9,729,868  & 30      & 202,538 (18.0\%)  \\ 
    \bottomrule
    \end{tabular}%
    }
  \label{tab_statistics}%
\end{table}%

\subsection{Baseline Methods}
To demonstrate the superiority of integrating LICAP with existing NIE methods and to also compare them to baseline methods, we introduce three sets of methods. All the three sets of methods are employed as the baselines for comparison, while GNN-based supervised methods are also integrated with LICAP to show the usefulness of LICAP.
\begin{itemize}
\item \textbf{Unsupervised Methods.}
PageRank (PR) \cite{page1999pagerank} and Personalised PageRank (PPR) \cite{haveliwala2002topic} are two representative algorithms to infer node importance without using labels for training. They mainly use random walks to capture graph topology for node importance estimation, and cannot utilize node features.

\item \textbf{Non-GNN Supervised Methods.}
Linear Regression (LR) and MultiLayer Perception (MLP) are two classical machine learning models, and can be applied to the node importance estimation problem as it is indeed a regression problem. We train LR and MLP regressors using the available node importance labels and corresponding node features. The trained regressor can be then used to predict node importance. %They are non-GNN models and cannot directly accept a graph as the input.

\item \textbf{GNN-based Supervised Methods.} There are two subsets of GNN-based supervised methods, and these methods can utilize both graph topology and node features at the same time. The first subset is the \textbf{general-purpose} GNN models, which are modified for node importance estimation problems. We choose GCN \cite{kipf2017semi} as the representative spectral GNN model, GraphSAGE \cite{hamilton2017inductive} as the representative spatial GNN model, and RGCN \cite{schlichtkrull2018modeling} as the representative GNN model for the relational graph data. The second subset is the \textbf{NIE-specific} GNN models. We employ GENI \cite{park2019estimating} and RGTN \cite{huang2021representation}, which achieve the state-of-the-art performance of node importance estimation on knowledge graphs, for comparison and benchmark.
\end{itemize}

\subsection{Evaluation Metrics}
In the evaluation of node importance estimation, both importance score prediction and importance score ranking are considered. As such, we leverage both regression and ranking metrics, which are formally defined as follows.

\begin{itemize}
\item \textbf{Regression Metrics.} Since the ground truth labels for node importance estimation problem are non-negative real numbers, it is natural to use regression metrics for evaluation. We utilize two commonly used regression metrics, namely Root Mean Square Error (RMSE) and Median Absolute Error (MedianAE), to quantify how close the predicted importance scores are to their corresponding ground truth labels. Lower values indicate better performance.

\textbf{RMSE} measures the average distance between the predicted scores and the ground truth scores. Considering $n$ samples and letting $s^*$ and $s$ denote predicted and actual scores respectively, we formally have:
\begin{equation}
\text{RMSE}=\sqrt{\frac{\sum_{i=1}^n (s^*_i-s_{i})^2}{n}}
\end{equation}

\textbf{MedianAE} measures the median of the absolute differences between the predicted scores and the ground truth scores. It is more robust to outliers than RMSE, since it focuses on the median prediction error rather than the overall errors of all samples. The formal definition of MedianAE is as follows:
\begin{equation}
\text{MedianAE}=\text{median}(\lvert s^*_1 - s_{1} \rvert, \cdots, \lvert s^*_n - s_{n} \rvert)
\end{equation}

\item \textbf{Ranking Metrics.} To evaluate the performance from another perspective, we follow \cite{huang2021representation} and additionally employ two ranking metrics: Normalized Discounted Cumulative Gain (NDCG) and Spearman’s rank correlation coefficient (SPEARMAN). Higher values indicate better performance.

\textbf{NDCG} measures the quality of a ranked list of items, which is an extension of Discounted Cumulative Gain (DCG). DCG takes into account both the relevance of the items and their position in the ranking. After ordering a list of items with ground truth $s$ by their corresponding predicted scores $s^*$, let $s_r$ denote the ground truth of the $r$-th item. DCG at position $k$ (DCG@k) \cite{huang2021representation} can be defined by:
\begin{equation}
\text{DCG@k}=\sum_{r=1}^k \frac{s_{r}} {\log_{2} (r+1)}
\end{equation}

NDCG further normalizes DCG by dividing it with the Ideal DCG (IDCG). The IDCG at positron $k$ (IDCG@k) is the maximum possible DCG among the top $k$ positions. With these definitions, NDCG at position $k$ (NDCG@k) can be then computed via:
\begin{equation}
\text{NDCG@k}=\frac{\text{DCG@k}} {\text{IDCG@k}}
\end{equation}

\textbf{SPEARMAN} is a measure of the rank correlation coefficient between the predicted scores $s^*$ and the ground-truth scores $s$. After converting scores $s^*$ and $s$ into the corresponding ranks $r^*$ and $r$ respectively, we define the mean of $r^*$ and $r$ as $\bar {r^*}$ and $\bar r$. SPEARMAN can be then formulated as:
\begin{equation}
\text{SPEARMAN}=\frac{\sum_{r} (r^*-\bar {r^*})(r-\bar r)} {\sqrt{\sum_{r} (r^*-\bar {r^*})^2} \sqrt{\sum_{r} (r-\bar r)^2}}
\end{equation}

\textbf{OVER} is the overlap ratio between the predicted important nodes and the real important nodes. Let $S^*_{top-k}$ denote the set of top $k$ important predicted nodes and $S_{top-k}$ denotes the set of top $k$ important ground-truth nodes. $\lvert \cdot \rvert$ denotes the cardinality of a set. OVER@k is calculated by:
\begin{equation}
\text{OVER@k}=\frac{\lvert S^*_{top-k} \cap S_{top-k} \rvert}{k}
\end{equation}

\end{itemize}

\subsection{Implementation Details}
We first employ LICAP to generate pretrained embeddings for each dataset. There pretrained embeddings are then fed to the downstream NIE models (if applicable, e.g., GNN-based models) respectively to verity the effectiveness of integrating with LICAP. All the NIE methods are implemented using Python. Specifically, PR and PPR are adapted from NetworkX package\footnote{\url{https://networkx.org}}; LR and MLP are adapted from scikit-learn package\footnote{\url{https://scikit-learn.org}}; GCN, GraphSAGE, and RGCN are adapted from DGL package\footnote{\url{https://www.dgl.ai}} with 1 hidden layer of 64 hidden dimensions. Regarding the state-of-the-art NIE methods GENI and RGTN, we use the implementations from RGTN \cite{huang2021representation} official GitHub repository\footnote{\url{https://github.com/GRAPH-0/RGTN-NIE}}. For LICAP, we search important ratio $\gamma$ [0.01, 0.05, 0.1, 0.2] and loss balancing ratio $\eta_2$ [0.5, 1, 10, 50] with $\eta_1$ being 1; we set temperature parameter $\tau$ of InfoNCE to 0.05. For PreGAT or GAT adopted in LICAP, we follow \cite{velivckovic2017gat} by setting hidden dimensions and number of heads to 8 respectively, and the dimensionality of predicate in PreGAT is set to 10. To make the comparison fair, we search learning rate [0.0005, 0.001, 0.005, 0.01] for both LICAP enhanced NIE methods and original NIE methods. 
%pretrain patience: 10 20 50 搜索；max epoch with early stopping  ~ 500, 1000 

As for datasets, we follow \cite{park2019estimating} and \cite{huang2021representation} using Node2Vec \cite{grover2016node2vec} to extract the structural features for each node given an input graph. Following \cite{huang2021representation}, the semantic features of FB15K and TMDB5K are both cloned from RGTN GitHub repository, while the semantic features of new dataset GA16K is obtained by running Transformer-XL \cite{dai2019TransformerXL}. The dimensions of pretrained structural features and semantic features are set to 64 and 128. The lengths of grouping bin for FB15K, TMDB5K, and GA16K are set to 1.0, 0.5, and 1.0. Regarding evaluation, five-fold cross validation is adopted \cite{park2019estimating,huang2021representation} and we report both regression metrics and ranking metrics.
%Node2Vec by pytorch geometri
%Transformer-XL by huggingface
%As for LICAP pretrained vectors, structural dim=64, semantic dim=128 (output dim for gat or pregat)
%semantic features: TransformerXL, 768 for tmdb and fb15k, 1024 for ga16k 为什么GA16K大于其他两个数据集？huggingface上只有1024
%five-fold cross validation 两个工作GENI和RGTN都用这个吗？

\section{Experiments and Discussion}
\begin{table*}[hbp]
  \centering
  \caption{Experimental results of different NIE methods over three real-world datasets and five metrics. Lower is better for regression metrics; higher is better for ranking metrics. The best performance is \textbf{bolded} and the runner-up is \underline{underlined}.}
    \begin{tabular}{llllll}
    \toprule
    \multirow{2}[4]{*}{Methods} & \multicolumn{5}{c}{FB15K} \\
\cmidrule{2-6}          & RMSE~$\downarrow$ & MedianAE~$\downarrow$ & NDCG@100~$\uparrow$ & SPEARMAN~$\uparrow$ & OVER@100~$\uparrow$ \\
    \midrule
    PR    & 9.6920±0.2006 & 9.6025±0.0468 & 0.8400±0.0103 & 0.3497±0.0188 & 0.1520±0.0172 \\
    PPR   & 9.6905±0.2041 & 9.5948±0.0474 & 0.8411±0.0112 & 0.3500±0.0192 & 0.1540±0.0185 \\
    \midrule
    LR    & 1.3115±0.0148 & 0.8156±0.0058 & 0.8663±0.0077 & 0.4771±0.0129 & 0.1580±0.0133 \\
    MLP   & 1.2074±0.0230 & 0.7286±0.0180 & 0.8960±0.0072 & 0.6072±0.0209 & 0.1920±0.0248 \\
    \midrule
    GCN   & 1.2236±0.0327 & 0.7210±0.0222 & 0.9336±0.0076 & 0.6142±0.0469 & 0.3560±0.0615 \\
    RGCN  & 1.3176±0.0476 & 0.6905±0.0232 & 0.8967±0.0171 & 0.6739±0.0248 & 0.3220±0.0703 \\
    GraphSAGE & 1.0810±0.0081 & 0.6510±0.0193 & 0.9047±0.0164 & 0.6866±0.0108 & 0.2640±0.0258 \\
    GENI  & 0.9868±0.0106 & 0.5808±0.0140 & 0.9222±0.0074 & 0.7405±0.0385 & 0.3300±0.0228 \\
    RGTN  & \underline{0.9343±0.0384} & \underline{0.5645±0.0391} & \underline{0.9408±0.0100} & \underline{0.7794±0.0100} & \underline{0.4000±0.0562} \\
    \midrule
    \textbf{RGTN+LICAP} & \textbf{0.8921±0.0290} & \textbf{0.5195±0.0072} & \textbf{0.9487±0.0063} & \textbf{0.7899±0.0068} & \textbf{0.4280±0.0564} \\
    \midrule
    \multirow{2}[4]{*}{Methods} & \multicolumn{5}{c}{TMDB5K} \\
\cmidrule{2-6}          & RMSE~$\downarrow$ & MedianAE~$\downarrow$ & NDCG@100~$\uparrow$ & SPEARMAN~$\uparrow$ & OVER@100~$\uparrow$ \\
    \midrule
    PR    & 2.1411±0.0114 & 2.0025±0.0371 & 0.8387±0.0102 & 0.6247±0.0130 & 0.4100±0.0616 \\
    PPR   & 1.8538±0.0101 & 1.6869±0.0241 & 0.8495±0.0083 & 0.6856±0.0098 & 0.4200±0.0506 \\
    \midrule
    LR    & 0.7672±0.0123 & 0.5124±0.0222 & 0.8843±0.0166 & 0.7370±0.0163 & 0.4820±0.0376 \\
    MLP   & 0.9054±0.0282 & 0.5888±0.0093 & 0.8525±0.0121 & 0.6568±0.0260 & 0.4360±0.0196 \\
    \midrule
    GCN   & 0.7925±0.0103 & 0.5419±0.0112 & 0.8752±0.0162 & 0.7553±0.0130 & 0.4920±0.0264 \\
    RGCN  & 0.7791±0.0182 & 0.5247±0.0228 & 0.8773±0.0202 & 0.7501±0.0164 & 0.5160±0.0242 \\
    GraphSAGE & 0.7327±0.0104 & 0.4794±0.0194 & 0.9020±0.0075 & 0.7635±0.0146 & 0.5400±0.0303 \\
    GENI  & 0.7369±0.0612 & 0.4894±0.0373 & 0.9136±0.0144 & 0.7823±0.0196 & 0.5560±0.0338 \\
    RGTN  & \underline{0.7080±0.0220} & \underline{0.4518±0.0329} & \underline{0.9159±0.0111} & \underline{0.7936±0.0112} & \underline{0.5720±0.0248} \\
    \midrule
    \textbf{RGTN+LICAP} & \textbf{0.6846±0.0084} & \textbf{0.4398±0.0169} & \textbf{0.9171±0.0099} & \textbf{0.7969±0.0111} & \textbf{0.5780±0.0331} \\
    \midrule
    \multirow{2}[4]{*}{Methods} & \multicolumn{5}{c}{GA16K} \\
\cmidrule{2-6}          & RMSE~$\downarrow$ & MedianAE~$\downarrow$ & NDCG@100~$\uparrow$ & SPEARMAN~$\uparrow$ & OVER@100~$\uparrow$ \\
    \midrule
    PR    & 5.3614±0.0251 & 5.6469±0.0203 & 0.8338±0.0160 & 0.7654±0.0055 & 0.2360±0.0196 \\
    PPR   & 5.0686±0.0235 & 5.3247±0.0207 & 0.8338±0.0160 & 0.7656±0.0055 & 0.2360±0.0196 \\
    \midrule
    LR    & 2.0679±0.0406 & 1.3233±0.0281 & 0.8055±0.0205 & 0.2303±0.0104 & 0.1740±0.0196 \\
    MLP   & 2.0118±0.0235 & 1.3374±0.0302 & 0.8124±0.0173 & 0.5193±0.0112 & 0.2060±0.0185 \\
    \midrule
    GCN   & 1.7133±0.0430 & 1.0855±0.0312 & 0.8773±0.0155 & 0.7986±0.0166 & 0.3420±0.0479 \\
    RGCN  & 1.2663±0.0275 & 0.6837±0.0147 & 0.8769±0.0194 & 0.8176±0.0100 & 0.3680±0.0337 \\
    GraphSAGE & 1.4055±0.0232 & 0.7913±0.0257 & 0.8140±0.0178 & 0.7602±0.0144 & 0.2140±0.0476 \\
    GENI  & 1.1944±0.0198 & 0.6442±0.0197 & 0.8326±0.0160 & 0.8242±0.0084 & 0.2520±0.0343 \\
    RGTN  & \underline{1.0608±0.0263} & \underline{0.5230±0.0221} & \underline{0.8799±0.0060} & \underline{0.8525±0.0122} & \underline{0.3840±0.0233} \\
    \midrule
    \textbf{RGTN+LICAP} & \textbf{1.0482±0.0158} & \textbf{0.5125±0.0252} & \textbf{0.8882±0.0073} & \textbf{0.8605±0.0093} & \textbf{0.3960±0.0233} \\
    \bottomrule
    \end{tabular}%
  \label{tab:main}%
\end{table*}%

To demonstrate the effectiveness of LICAP and its components, we investigate the following research questions.
\begin{itemize}
\item Can LICAP further boost the performance of the state-of-the-art NIE method and achieve the new state-of-the-art performance? (Section \ref{exp:main})
\item Do the top nodes preferred hierarchical sampling and two contrastive learning losses work? (Section \ref{exp:loss})
\item Is the newly proposed PreGAT, which considers predicates for GAT, better than GAT as the pretraining model of LICAP? (Section \ref{exp:PreGAT})
\item How are the performances of LICAP to pretrain structural features, semantic features, and the both features respectively? (Section \ref{exp:Channels})
\item How do the key hyper-parameters of LICAP affect the performance? (Section \ref{exp:parameter})
\item Can the LICAP pretrained embeddings be better aware of the nodes with high importance scores? (Section \ref{exp:vis})
\item Can LICAP further improve the performance of other NIE methods? (Section \ref{exp:plug-in})
\end{itemize}

\subsection{Node Importance Estimation on Knowledge Graphs} \label{exp:main}
We illustrate the performance comparison of the proposed LICAP and several baseline methods in TABLE \ref{tab:main}. The main observations from the table are as follows.

%思路：4小段 每段首句为主旨句。
% 1 LICAP 2 PR/PPR/LR/MLP 3 GNN METHODS 4 RGTN
% 顺序：1234 or 2341 ，前者强调LICAP 后者强调模型顺序
(1) The proposed LICAP integrated with RGTN consistently outperforms all baseline methods over all metrics. In particular, LICAP gains significant improvements on regression metrics comparing with vanilla RGTN, which can be owing to the label informed contrastive pretraining mechanism. In fact, LICAP can also enhance the performance of other NIE models to varying degrees, which is discussed in great details in Section \ref{exp:plug-in}.

%(2) Compared to other supervised methods, unsupervised methods like PR and PPR have poor performance on RMSE and MedianAE. This is because unsupervised methods cannot utilize the supervised signals, thus resulting in poor regression metrics.
(2) The unsupervised methods, i.e., PR and PPR, obtain poor performance comparing to the supervised methods. This is because the unsupervised methods cannot use the available labels for supervision, thus rendering poor performance.

(3) The GNN-based methods receive superior performance comparing to the non-GNN based methods in most cases. Regarding the non-GNN methods, PR and PPR fail to use node features and available labels; LR and MLP cannot naturally include graph structure. However, GNN-based methods can utilize both node features and graph structure, thus resulting in superior performance. 
%In partitular, GNN-based methods considering relational data have better results, such as RGCN and GENI. 

(4) RGTN performs better than all the other baseline methods except the LICAP enhanced RGTN version for all regression and ranking metrics. There could be two possible reasons. First, the relative ranking information among nodes are employed in RGTN learning process as an auxiliary loss. Second, RGTN explicitly considers semantic features and structural features through two channels, which also motivates us to conduct further experiments in Section \ref{exp:Channels}.

\subsection{Ablation Study: Effect of LICAP Contrastive Losses} \label{exp:loss}
\begin{table*}[hbp]
  \centering
  \caption{LICAP loss: ablation study with LICAP variants based on RGTN}
    \begin{tabular}{llllll}
    \toprule
    \multirow{2}[4]{*}{Methods} & \multicolumn{5}{c}{FB15K} \\
\cmidrule{2-6}          & RMSE~$\downarrow$ & MedianAE~$\downarrow$ & NDCG@100~$\uparrow$ & SPEARMAN~$\uparrow$ & OVER@100~$\uparrow$ \\
    \midrule
    Vanilla & 0.9343±0.0384 & 0.5645±0.0391 & 0.9408±0.0100 & 0.7794±0.0100 & 0.4000±0.0562 \\
    +LICAP-r.d. & 0.9408±0.0507 & 0.5746±0.0545 & 0.9409±0.0057 & 0.7745±0.0113 & 0.3940±0.0450 \\
    +LICAP-$L_1$ & \underline{0.9210±0.0341} & \underline{0.5337±0.0164} & \underline{0.9436±0.0114} & \textbf{0.7921±0.0088} & \underline{0.4020±0.0614} \\
    +LICAP-$L_2$ & 0.9598±0.0343 & 0.5745±0.0434 & 0.9354±0.0114 & 0.7658±0.0183 & 0.3860±0.0500 \\
    \textbf{+LICAP-$L_1$+$L_2$} & \textbf{0.8921±0.0290} & \textbf{0.5195±0.0072} & \textbf{0.9487±0.0063} & \underline{0.7899±0.0068} & \textbf{0.4280±0.0564} \\
    \midrule
    \multirow{2}[4]{*}{Methods} & \multicolumn{5}{c}{TMDB5K} \\
\cmidrule{2-6}          & RMSE~$\downarrow$ & MedianAE~$\downarrow$ & NDCG@100~$\uparrow$ & SPEARMAN~$\uparrow$ & OVER@100~$\uparrow$ \\
    \midrule
    Vanilla & 0.7080±0.0220 & 0.4518±0.0329 & \underline{0.9159±0.0111} & 0.7936±0.0112 & 0.5720±0.0248 \\
    +LICAP-r.d. & 0.7127±0.0356 & 0.4621±0.0298 &  0.9104±0.0077 &  0.7925±0.0098 &  0.5620±0.0371 \\
    +LICAP-$L_1$ & \underline{0.6904±0.0172} & \textbf{0.4369±0.0237} & 0.9129±0.0089 & \underline{0.7942±0.0100} & \underline{0.5740±0.0233} \\
    +LICAP-$L_2$ & 0.7156±0.0411 & 0.4844±0.0487 & 0.9054±0.0130 & 0.7800±0.0239 & 0.5520±0.0366 \\
    \textbf{+LICAP-$L_1$+$L_2$} & \textbf{0.6846±0.0084} & \underline{0.4398±0.0169} & \textbf{0.9171±0.0099} & \textbf{0.7969±0.0111} & \textbf{0.5780±0.0331} \\
    \midrule
    \multirow{2}[4]{*}{Methods} & \multicolumn{5}{c}{GA16K} \\
\cmidrule{2-6}          & RMSE~$\downarrow$ & MedianAE~$\downarrow$ & NDCG@100~$\uparrow$ & SPEARMAN~$\uparrow$ & OVER@100~$\uparrow$ \\
    \midrule
    Vanilla & 1.0608±0.0263 & 0.5230±0.0221 & 0.8799±0.0060 & 0.8525±0.0122 & 0.3840±0.0233 \\
    +LICAP-r.d. & 1.0706±0.0099 & 0.5312±0.0171 &  0.8800±0.0076 & 0.8518±0.0104 & 0.3860±0.0233 \\
    +LICAP-$L_1$ & 1.0589±0.0346 & 0.5264±0.0288 & \underline{0.8807±0.0182}  & \underline{0.8546±0.0064} & \underline{0.3900±0.0352} \\
    +LICAP-$L_2$ & \underline{1.0564±0.0214} & \underline{0.5211±0.0192} & 0.8756±0.0075 & 0.8527±0.0086 & 0.3600±0.0276 \\
    \textbf{+LICAP-$L_1$+$L_2$} & \textbf{1.0482±0.0158} & \textbf{0.5125±0.0252} & \textbf{0.8882±0.0073} & \textbf{0.8605±0.0093} & \textbf{0.3960±0.0233} \\
    \bottomrule
    \end{tabular}%
  \label{tab:loss}%
\end{table*}%

To investigate the impact of the two contrastive losses (corresponding to the two sets of contrastive samples via the proposed top nodes preferred hierarchical sampling) in LICAP, we compare the variants of LICAP on the basis of RGTN.
\begin{itemize}
\item Vanilla denotes the original RGTN method. 
\item LICAP-r.d. contains $L_1$ with random sampling.
\item LICAP-$L_1$ contains only $L_1$ loss with unit weight.
\item LICAP-$L_2$ contains only $L_2$ loss with unit weight.
\item LICAP-$L_1$+$L_2$ contains both $L_1$ and $L_2$ with the loss balancing weights as used in Section \ref{exp:main}.
\end{itemize}

As shown in TABLE \ref{tab:loss}, we can observe that the results of LICAP-$L_1$+$L_2$ are the best among these variants, which confirms the effectiveness of our two proposed contrastive losses. Besides, it is interesting to find that LICAP-$L_1$ mostly performs better than LICAP-$L_2$. We speculate that applying $L_2$ alone is not enough for label informed contrastive learning, since $L_2$ alone directly looks at too finer details and makes the learning process difficult. This is also the reason why we need the so-called hierarchical sampling and jointly optimizing $L_1$ (more global view) together with $L_2$.

To further demonstrate the usefulness of the top nodes preferred sampling strategy, we randomly choose the equal number of nodes as the number of top nodes chosen by LICAP-$L_1$ (or LICAP-$L_2$), and term it as LICAP-r.d. method. Specifically, for positive contrast, LICAP-r.d. randomly selects nodes (rather than top nodes) and contrast them positively to their element-wisely averaging prototype; for negative contrast, LICAP-r.d. randomly selects from the residual nodes as the negative node to each positive node. According to the results in Table III, LICAP-$L_1$ consistently outperforms LICAP-r.d. This observation is in accord with our motivation that injecting the prior knowledge of being better aware of top nodes can help NIE method further improve the performance.

\subsection{Ablation Study: Effect of Predicate-aware GAT} \label{exp:PreGAT}
To validate the effectiveness of the proposed PreGAT, we compare LICAP-PreGAT to the vanilla RGTN and LICAP-GAT. Concretely, LICAP-PreGAT and LICAP-GAT refer to integrating PreGAT and GAT to RGTN respectively. According to TABLE \ref{tab:gat}, LICAP-PreGAT consistently outperforms LICAP-GAT, and they both outperform the vanilla version. These observations imply not only the usefulness of LICAP framework, but also the further performance gains using PreGAT by additionally exploiting the predicates (i.e., relational information) of knowledge graphs.

To explore the relationship between dataset characteristics and the effectiveness of PreGAT, we observe from TABLE \ref{tab:gat} that PreGAT outperforms GAT most significantly on FB15K, whose number of predicates are extremely large as shown in Table~\ref{tab_statistics}. According to the observation that PreGAT works well with the graph with abundant predicates, we may conclude that the predicate-aware mechanism can differentiate among the diverse predicates and capture meaningful information among them. Contrarily, considering an extreme example of a graph with only one predicate (i.e., all edges are in the same type), the predicate embeddings would be same for all edges, and PreGAT would thus degrade to GAT. As a result, the gain of PreGAT over the gain of GAT is likely to be relatively low for the graphs with fewer predicates such as TMDB5K and GA16K. With above analysis, we further clarify the contributions of PreGAT for knowledge graphs.
\begin{table*}[htbp]
  \centering
  \caption{PreGAT or GAT: ablation study with LICAP variants based on RGTN.}
    \begin{tabular}{llllll}
    \toprule
    \multirow{2}[4]{*}{Methods} & \multicolumn{5}{c}{FB15K} \\
\cmidrule{2-6}          & RMSE~$\downarrow$ & MedianAE~$\downarrow$ & NDCG@100~$\uparrow$ & SPEARMAN~$\uparrow$ & OVER@100~$\uparrow$ \\
    \midrule
    Vanilla &  0.9343±0.0384  &  0.5645±0.0391 & 0.9408±0.0100  &  0.7794±0.0100  &  0.4000±0.0562  \\
    +LICAP-GAT &  \underline{0.9057±0.0228}  &  \underline{0.5285±0.0184} & \underline{0.9439±0.0138}  &  \underline{0.7844±0.0118}  &  \underline{0.4100±0.0490}  \\
    \textbf{+LICAP-PreGAT} & \textbf{0.8921±0.0290} & \textbf{0.5195±0.0072} & \textbf{0.9487±0.0063 } & \textbf{0.7899±0.0068} & \textbf{0.4280±0.0564} \\
    \midrule
    \multirow{2}[4]{*}{Methods} & \multicolumn{5}{c}{TMDB5K} \\
\cmidrule{2-6}          & RMSE~$\downarrow$ & MedianAE~$\downarrow$ & NDCG@100~$\uparrow$ & SPEARMAN~$\uparrow$ & OVER@100~$\uparrow$ \\
    \midrule
    Vanilla &  0.7080±0.0220  &  0.4518±0.0329 & \underline{0.9159±0.0111}  &  0.7936±0.0112  & 0.5720±0.0248 \\
    +LICAP-GAT &  \underline{0.6895±0.0042}  &  \underline{0.4439±0.0116} & 0.9141±0.0138  &  \underline{0.7955±0.0099}  &  \underline{0.5720±0.0194}  \\
    \textbf{+LICAP-PreGAT} & \textbf{0.6846±0.0084} & \textbf{0.4398±0.0169} & \textbf{0.9171±0.0099} & \textbf{0.7969±0.0111} & \textbf{0.5780±0.0331} \\
    \midrule
    \multirow{2}[4]{*}{Methods} & \multicolumn{5}{c}{GA16K} \\
\cmidrule{2-6}          & RMSE~$\downarrow$ & MedianAE~$\downarrow$ & NDCG@100~$\uparrow$ & SPEARMAN~$\uparrow$ & OVER@100~$\uparrow$ \\
    \midrule
    Vanilla &  1.0608±0.0263  &  0.5230±0.0221 & \underline{0.8799±0.0060}  &  0.8525±0.0122  &  0.3840±0.0233  \\
    +LICAP-GAT &  \underline{1.0559±0.0175}  &  \underline{0.5156±0.0315} & 0.8740±0.0095  &  \underline{0.8543±0.0039}  &  \underline{0.3940±0.0136}  \\
    \textbf{+LICAP-PreGAT} & \textbf{1.0482±0.0158} & \textbf{0.5125±0.0252} & \textbf{0.8882±0.0073} & \textbf{0.8605±0.0093} & \textbf{0.3960±0.0233} \\
    \bottomrule
    \end{tabular}%
  \label{tab:gat}%
\end{table*}%

\subsection{Pretraining Different Input Channels} \label{exp:Channels}
The input to GNN models could be structural features, semantic features, or both features. In particular, the current best NIE model for knowledge graphs is RGTN that explicitly considers structural features and semantic features as two individual channels. To explore the differences of using LICAP over different combinations of channel(s), we implement and compare the following LICAP variants:
\begin{itemize}
\item Vanilla denotes the original RGTN method. 
\item LICAP (structural) only pretrains structural embeddings for downstream RGTN.
\item LICAP (semantic) only pretrains semantic embeddings for downstream RGTN.
\item LICAP (both) pretrains both structural and semantic embeddings for downstream RGTN.
\end{itemize}

TABLE \ref{tab:rgtnchannel} shows that LICAP (structural) and LICAP (semantic) often obtain superior results than LICAP (both). A possible reason is that the input embeddings of RGTN's two channels are supposed to be distinct to a certain extent, which is beneficial for utilizing multiple-sources features. Thus, pretraining both structural and semantic embeddings following the same LICAP architecture may harm the model's learning ability. 
Nevertheless, we can always obtain the best performance with one of the three LICAP variants compared to the vanilla version.
\begin{table*}[htbp]
  \centering
  \caption{The ablation of LICAP variants on different RGTN channels.}
    \begin{tabular}{llllll}
    \toprule
    \multirow{2}[4]{*}{Methods} & \multicolumn{5}{c}{FB15K} \\
\cmidrule{2-6}          & RMSE~$\downarrow$ & MedianAE~$\downarrow$ & NDCG@100~$\uparrow$ & SPEARMAN~$\uparrow$ & OVER@100~$\uparrow$ \\
    \midrule
    Vanilla  & 0.9343±0.0384 & 0.5645±0.0391 & 0.9408±0.0100 & 0.7794±0.0100 & 0.4000±0.0562 \\
    \textbf{+LICAP} (structural) & \underline{0.8921±0.0371} & \underline{0.5265±0.0272} & \underline{0.9469±0.0057} & \textbf{0.7916±0.0103} & \textbf{0.4300±0.0555} \\
    \textbf{+LICAP} (semantic) & \textbf{0.8921±0.0290} & \textbf{0.5195±0.0072} & \textbf{0.9487±0.0063} & \underline{0.7899±0.0068} & \underline{0.4280±0.0564} \\
    \textbf{+LICAP} (both) & 0.9153±0.0280 & 0.5418±0.0187 & 0.9452±0.0097 & 0.7816±0.0092 & 0.4140±0.0683 \\
    \midrule
    \multirow{2}[4]{*}{Methods} & \multicolumn{5}{c}{TMDB5K} \\
\cmidrule{2-6}          & RMSE~$\downarrow$ & MedianAE~$\downarrow$ & NDCG@100~$\uparrow$ & SPEARMAN~$\uparrow$ & OVER@100~$\uparrow$ \\
    \midrule
    Vanilla  & 0.7080±0.0220 & 0.4518±0.0329 & \underline{0.9159±0.0111} & 0.7936±0.0112 & \underline{0.5720±0.0248} \\
    \textbf{+LICAP} (structural) & \underline{0.6967±0.0142} & 0.4426±0.0143 & 0.9123±0.0114 & \textbf{0.7977±0.0029} & 0.5620±0.0286 \\
    \textbf{+LICAP} (semantic) & \textbf{0.6846±0.0084} & \textbf{0.4398±0.0169} & \textbf{0.9171±0.0099} & \underline{0.7969±0.0111} & \textbf{0.5780±0.0331} \\
    \textbf{+LICAP} (both) & 0.6985±0.0130 & \underline{0.4418±0.0154} & 0.9077±0.0066 & 0.7888±0.0104 & 0.5580±0.0564 \\
    \midrule
    \multirow{2}[4]{*}{Methods} & \multicolumn{5}{c}{GA16K} \\
\cmidrule{2-6}          & RMSE~$\downarrow$ & MedianAE~$\downarrow$ & NDCG@100~$\uparrow$ & SPEARMAN~$\uparrow$ & OVER@100~$\uparrow$ \\
    \midrule
    Vanilla  & 1.0608±0.0263 & 0.5230±0.0221 & \underline{0.8799±0.0060} & 0.8525±0.0122 & \underline{0.3840±0.0233} \\
    \textbf{+LICAP} (structural) & \textbf{1.0395±0.0177} & \textbf{0.5026±0.0183} & 0.8729±0.0170 & \underline{0.8551±0.0062} & 0.3800±0.0200 \\
    \textbf{+LICAP} (semantic) & \underline{1.0482±0.0158} & 0.5125±0.0252 & \textbf{0.8882±0.0073} & \textbf{0.8605±0.0093} & \textbf{0.3960±0.0233} \\
    \textbf{+LICAP} (both) & 1.0564±0.0202 & \underline{0.5119±0.0176} & 0.8685±0.0169 & 0.8222±0.0102 & 0.3600±0.0322 \\
    \bottomrule
    \end{tabular}%
  \label{tab:rgtnchannel}%
\end{table*}%

\subsection{Hyper-parameter Analysis} \label{exp:parameter}
There are two key hyper-parameters for LICAP, i.e., loss balancing ratio $\eta_2/\eta_1$ and important ratio $\gamma$. RMSE and SPEARMAN are used as the metrics for illustration, as they can reflect regression and ranking performances respectively. We vary $\gamma$ and $\eta_2/\eta_1$ in the range of 0.01 to 0.20 and 0.02 to 2.00 respectively while fixing other parameters.
\begin{figure}[htbp]
\centering
\includegraphics[width=\linewidth]{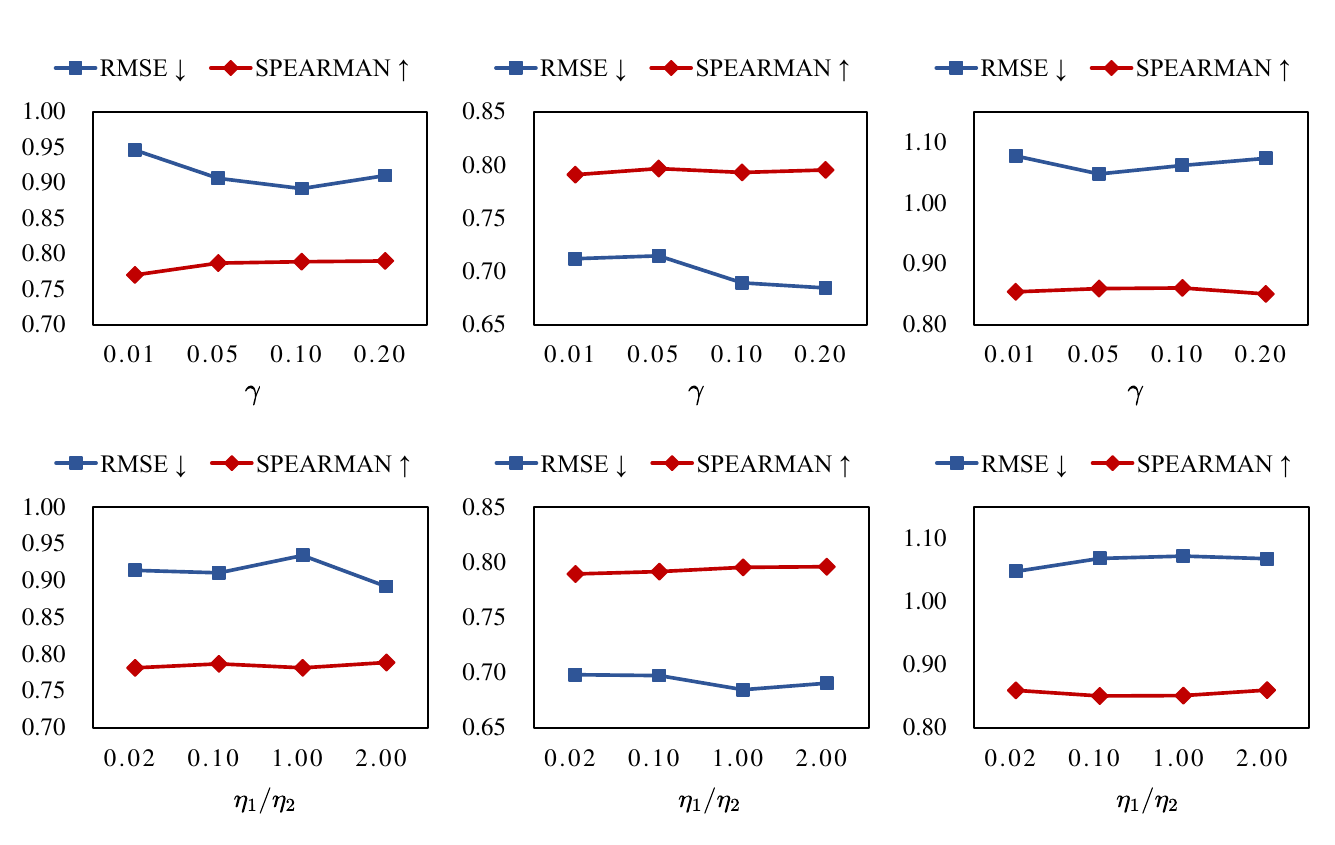}
\caption{Performance under different hyper-parameters of $\eta1/\eta2$ and $\gamma$ on FB15K (left), TMDB5K (middle) and GA16K (right).}
\label{fig:para}
\end{figure}

According to Fig.\ref{fig:para}, we find that SPEARMAN (ranking metric) is more robust to $\gamma$ and $\eta_2/\eta_1$ compared to RMSE (regression metric). The reason is that ranking is relatively less sensitive to predictive outliers, while regression especially using RMSE as the metric would be greatly affected by predictive outliers. Moreover, we also observe that the best performance occurs at different $\gamma$ for three datasets. This is because that the three datasets have different characteristics such as graph structure and KG theme.

\subsection{Visualizing LICAP Pretrained Embeddings} \label{exp:vis}
\begin{figure}[htbp]
\centering
\includegraphics[width=\linewidth]{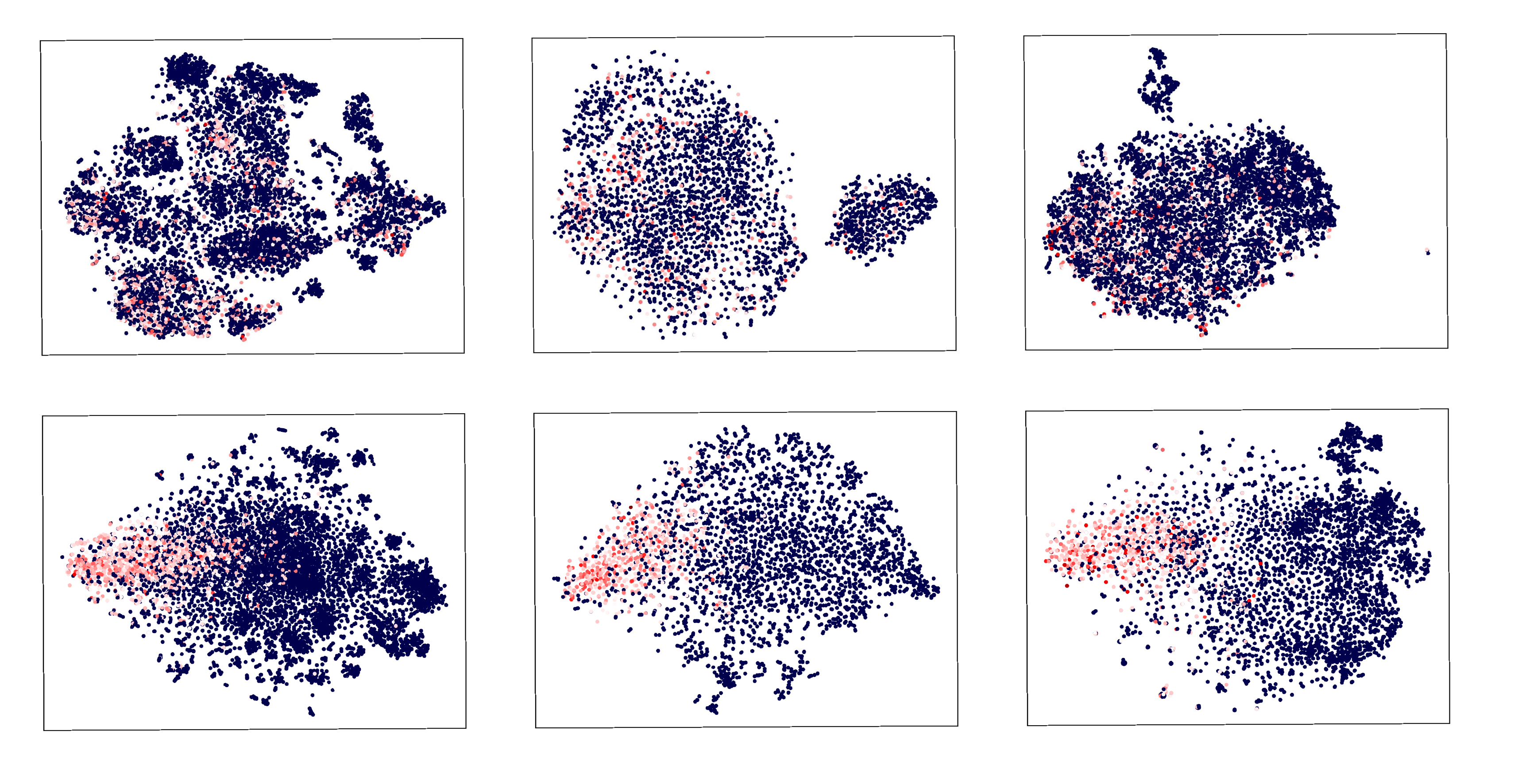}
    \caption{Visualization of unpretrained (upper) and LICAP pretrained (bottom) node embeddings on FB15K (left), TMDB5K (middle) and GA16K (right).}
\label{fig:vis}
\end{figure}

The main motivation of this work is to be better aware of the top nodes with high importance scores. To intuitively understand if the goal is achieved, we visualize the pretrained LICAP embeddings and their unpretrained counterparts in Fig.\ref{fig:vis}. Concretely, we choose PreGAT as the pretraining model and check the node semantic embeddings before and after pretraining. We run t-SNE \cite{van2008visualizing} and annotate top nodes and non-top nodes by red and blue respectively. It is easy to observe from Fig.\ref{fig:vis} that the top nodes after LICAP pretraining are more gathered on the one side of the space, which is the good indicator of being better aware of top nodes.

% \subsection{Analysis of Regression based Contrastive Learning}
% using embedding visualization (as well as Section \ref{exp:parameter}) to analyze Contrastive Loss and Sampling Strategy
%咋们这个方法需要取个名字，比如叫“Regression/Ranking based Contrastive Learning”
%通过删除模型中特定难度范围的负样本来分析这些样本对整个模型性能的影响

%\subsection{Case Study}?
%See GENI, the last page, C1 Case Study
%咋们既然更关注头部数据，那么如果把头部结果打印出来，能比别人头部结果好，
% 对着结果和~$\downarrow$场景~$\downarrow$做个解释，这种类型叫case study

%\subsection{Wall-clock Time}
%如果我们新增策略不会增加很多运行时间，可以增加这一节

\subsection{Applying LICAP to Various Methods} \label{exp:plug-in}
%Applying LICAP to other GNN-based methods-----
The proposed LICAP is introduced in pretraining stage, rendering it more flexibility. Therefore, we further investigate the effectiveness of LICAP combined with more NIE methods. The results are reported in TABLE \ref{tab:downstreamgnn}. 

Regarding non-GNN based methods, LICAP consistently improves the performance of MLP, while it may fail for LR at some metrics on TMDB5K and GA16K. We speculate that this is because of LR’s linear hypothesis being not compatible well for NIE problems on knowledge graphs, whereas MLP with non-linear activation can well collaborate with LICAP. 

Regarding GNN-based methods, we observe that LICAP improves the performance of GCN, RGCN, GraphSAGE and GENI to varying degrees. Among these NIE methods, RGCN and GENI integrated with LICAP have relative consistent better results across different datasets compared to their vanilla versions. The reason is that LICAP considers relational information, which is more suitable for the models also considering relations such as RGCN and GENI.  

Comparing the results of non-GNN and GNN based methods, we find that LICAP brings relatively higher performance gains for non-GNN methods, possibly because the PreGAT module in LICAP incorporates the topological information omitted in LR or MLP, making them more complementary to each other. Despite that, GNN-based methods with LICAP still achieve better results due to the superiority of GNN backbones comparing to non-GNN backbones on graph data.

\begin{table*}[htbp]
  \centering
  \caption{Performance comparison w.r.t. various downstream models.}
    \begin{tabular}{lllllll}
    \toprule
    \multirow{2}[4]{*}{} & \multirow{2}[4]{*}{Methods} & \multicolumn{5}{c}{FB15K} \\
\cmidrule{3-7}          &       & RMSE~$\downarrow$ & MedianAE~$\downarrow$ & NDCG@100~$\uparrow$ & SPEARMAN~$\uparrow$ & OVER@100~$\uparrow$ \\
    \midrule
    \multirow{4}[4]{*}{Non-GNN based} & LR    & 1.3115±0.0148 & 0.8156±0.0058 & 0.8663±0.0077 & 0.4771±0.0129 & 0.1580±0.0133 \\
          & \textbf{+LICAP} & \textbf{1.2622±0.0188} & \textbf{0.7635±0.0187} & \textbf{0.8842±0.0172} & \textbf{0.5484±0.0318} & \textbf{0.2080±0.0402} \\
\cmidrule{2-7}          & MLP   & 1.2074±0.0230 & 0.7286±0.0180 & 0.8960±0.0072 & 0.6072±0.0209 & 0.1920±0.0248 \\
          & \textbf{+LICAP} & \textbf{1.1752±0.0586} & \textbf{0.6956±0.0402} & \textbf{0.9100±0.0087} & \textbf{0.6302±0.0440} & \textbf{0.2800±0.0663} \\
    \midrule
    \multirow{8}[8]{*}{GNN based} & GCN   & 1.2236±0.0327 & 0.7210±0.0222 & 0.9336±0.0076 & 0.6142±0.0469 & 0.3560±0.0615 \\
          & \textbf{+LICAP} & \textbf{1.1687±0.0449} & \textbf{0.6784±0.0154} & \textbf{0.9340±0.0056} & \textbf{0.6628±0.0258} & \textbf{0.3640±0.0326} \\
\cmidrule{2-7}          & RGCN  & 1.3176±0.0476 & 0.6905±0.0232 & 0.8967±0.0171 & \textbf{0.6739±0.0248} & 0.3220±0.0703 \\
          & \textbf{+LICAP} & \textbf{1.3069±0.0729} & \textbf{0.6876±0.0285} & \textbf{0.9116±0.0156} & 0.6701±0.0348 & \textbf{0.3240±0.0791} \\
\cmidrule{2-7}          & GraphSAGE & \textbf{1.0810±0.0081} & 0.6510±0.0193 & \textbf{0.9047±0.0164} & \textbf{0.6866±0.0108} & 0.2640±0.0258 \\
          & \textbf{+LICAP} & 1.1196±0.0369 & \textbf{0.6480±0.0198} & 0.9006±0.0202 & 0.6702±0.0222 & \textbf{0.2800±0.0261} \\
\cmidrule{2-7}          & GENI  & 0.9868±0.0106 & 0.5808±0.0140 & 0.9222±0.0074 & 0.7405±0.0385 & 0.3300±0.0228 \\
          & \textbf{+LICAP} & \textbf{0.9503±0.0111} & \textbf{0.5510±0.0175} & \textbf{0.9289±0.0092} & \textbf{0.7567±0.0124} & \textbf{0.3680±0.0223} \\
    \midrule
    \multirow{2}[4]{*}{} & \multirow{2}[4]{*}{Methods} & \multicolumn{5}{c}{TMDB5K} \\
\cmidrule{3-7}          &       & RMSE~$\downarrow$ & MedianAE~$\downarrow$ & NDCG@100~$\uparrow$ & SPEARMAN~$\uparrow$ & OVER@100~$\uparrow$ \\
    \midrule
    \multirow{4}[4]{*}{Non-GNN based} & LR    & \textbf{0.7672±0.0123} & \textbf{0.5124±0.0222} & \textbf{0.8843±0.0166} & \textbf{0.7370±0.0163} & 0.4820±0.0376 \\
          & \textbf{+LICAP} & 0.8140±0.0538 & 0.5516±0.0522 & 0.8716±0.0236 & 0.7071±0.0413 & \textbf{0.4880±0.0337} \\
\cmidrule{2-7}          & MLP   & 0.9054±0.0282 & 0.5888±0.0093 & 0.8525±0.0121 & 0.6568±0.0260 & 0.4360±0.0196 \\
          & \textbf{+LICAP} & \textbf{ 0.8461±0.0548} & \textbf{0.5517±0.0475} & \textbf{0.8785±0.0166} & \textbf{0.6744±0.0537} & \textbf{0.4840±0.0459} \\
    \midrule
    \multirow{8}[8]{*}{GNN based} & GCN   & \textbf{0.7925±0.0103} & 0.5419±0.0112 & 0.8752±0.0162 & \textbf{0.7553±0.0130} & \textbf{0.4920±0.0264} \\
          & \textbf{+LICAP} & 0.7974±0.0204 & \textbf{0.5335±0.0182} & \textbf{0.8776±0.0105} & 0.7384±0.0094 & 0.4780±0.0435 \\
\cmidrule{2-7}          & RGCN  & 0.7791±0.0182 & 0.5247±0.0228 & 0.8773±0.0202 & 0.7501±0.0164 & \textbf{0.5160±0.0242} \\
          & \textbf{+LICAP} & \textbf{0.7448±0.0341} & \textbf{0.4874±0.0342} & \textbf{0.8829±0.0140} & \textbf{0.7617±0.0194} & 0.5000±0.0648 \\
\cmidrule{2-7}          & GraphSAGE & \textbf{0.7327±0.0104} & 0.4794±0.0194 & \textbf{0.9020±0.0075} & \textbf{0.7635±0.0146} & \textbf{0.5400±0.0303} \\
          & \textbf{+LICAP} & 0.7399±0.0398 & \textbf{0.4706±0.0160} & 0.8997±0.0122 & 0.7603±0.0168 & 0.5260±0.0492 \\
\cmidrule{2-7}          & GENI  & 0.7369±0.0612 & 0.4894±0.0373 & \textbf{0.9136±0.0144} & 0.7823±0.0196 & \textbf{0.5560±0.0338} \\
          & \textbf{+LICAP} & \textbf{0.6895±0.0144} & \textbf{0.4417±0.0146} & 0.9036±0.0107 & \textbf{0.7874±0.0092} & 0.5360±0.0388 \\
    \midrule
    \multirow{2}[4]{*}{} & \multirow{2}[4]{*}{Methods} & \multicolumn{5}{c}{GA16K} \\
\cmidrule{3-7}          &       & RMSE~$\downarrow$ & MedianAE~$\downarrow$ & NDCG@100~$\uparrow$ & SPEARMAN~$\uparrow$ & OVER@100~$\uparrow$ \\
    \midrule
    \multirow{4}[4]{*}{Non-GNN based} & LR    & 2.0679±0.0406 & 1.3233±0.0281 & \textbf{0.8055±0.0205} & 0.2303±0.0104 & \textbf{0.1740±0.0196} \\
          & \textbf{+LICAP} & \textbf{1.9018±0.0271} & \textbf{1.1464±0.0650} & 0.7645±0.0367 & \textbf{0.4137±0.0386} & 0.1720±0.0264 \\
\cmidrule{2-7}          & MLP   & 2.0118±0.0235 & 1.3374±0.0302 & 0.8124±0.0173 & 0.5193±0.0112 & 0.2060±0.0185 \\
          & \textbf{+LICAP} & \textbf{1.3934±0.0323} & \textbf{0.6755±0.0167} & \textbf{0.8157±0.0162} & \textbf{0.7450±0.0120} & \textbf{0.2620±0.0366} \\
    \midrule
    \multirow{8}[8]{*}{GNN based} & GCN   & \textbf{1.7133±0.0430} & 1.0855±0.0312 & 0.8773±0.0155 & \textbf{0.7986±0.0166} & \textbf{0.3420±0.0479} \\
          & \textbf{+LICAP} & 1.7287±0.0275 & \textbf{1.0524±0.0465} & \textbf{0.8782±0.0189} & 0.7505±0.0143 & 0.3400±0.0210 \\
\cmidrule{2-7}          & RGCN  & 1.2663±0.0275 & 0.6837±0.0147 & 0.8769±0.0194 & 0.8176±0.0100 & 0.3680±0.0337 \\
          & \textbf{+LICAP} & \textbf{1.1746±0.0315} & \textbf{0.5456±0.0185} & \textbf{0.8844±0.0105} & \textbf{0.8362±0.0081} & \textbf{0.3960±0.0301} \\
\cmidrule{2-7}          & GraphSAGE & 1.4055±0.0232 & 0.7913±0.0257 & 0.8140±0.0178 & 0.7602±0.0144 & 0.2140±0.0476 \\
          & \textbf{+LICAP} & \textbf{1.2594±0.0243} & \textbf{0.6222±0.0159} & \textbf{0.8472±0.0234} & \textbf{0.7991±0.0121} & \textbf{0.3200±0.0141} \\
\cmidrule{2-7}          & GENI  & \textbf{1.1944±0.0198} & 0.6442±0.0197 & 0.8326±0.0160 & \textbf{0.8242±0.0084} & 0.2520±0.0343 \\
          & \textbf{+LICAP} & 1.2213±0.0156 & \textbf{0.5838±0.0137} & \textbf{0.8670±0.0196} & 0.8204±0.0108 & \textbf{0.3420±0.0312} \\
    \bottomrule
    \end{tabular}%
  \label{tab:downstreamgnn}%
\end{table*}%

\section{Conclusion}
To be better aware of top nodes with high importance scores in knowledge graphs, this work proposed Label Information ContrAstive Pretraining (LICAP) to boost the performance of existing NIE methods. Specifically, LICAP was introduced in the pretraining stage in which a predicate-aware GAT was developed to pretrain embeddings. To fully utilize available node importance scores, the top nodes preferred hierarchical sampling was proposed, aiming to transform regression problem into classification-like problem, so as to incorporate with contrastive learning to guide PreGAT during pretraining.

We conducted a comprehensive empirical study on LICAP over nine NIE methods, three real-world knowledge graphs, and five regression or ranking metrics. The empirical results confirmed that applying LICAP to RGTN achieved the new state-of-the-art performance. In addition, LICAP was proved to broadly boost the performance of most existing NIE methods. And the last but not least, the usefulness of the components in LICAP such as top nodes preferred hierarchical sampling and PreGAT were also verified. For future work, it is interesting to extend LICAP to the NIE problem on dynamic graphs \cite{geng2022modeling}. Besides, it is also worth investigating how the label distribution affects the performance of LICAP and NIE methods. Another promising direction is to migrate the idea of LICAP to other graph-related regression or ranking problems.

\section*{Acknowledgments}
%This should be a simple paragraph before the References to thank those individuals and institutions who have supported your work on this article.
This work was funded by the Natural Science Foundation of China (No. 42050101) and the National Key R\&D Program of China (No. 2022YFF1202403). This is also a contribution to the IUGS Deep-time Digital Earth Big Science Program.

%{\appendix[Proof of the Zonklar Equations]
%Use $\backslash${\tt{appendix}} if you have a single appendix: Do not use $\backslash${\tt{section}} anymore after $\backslash${\tt{appendix}}, only $\backslash${\tt{section*}}. If you have multiple appendixes use $\backslash${\tt{appendices}} then use $\backslash${\tt{section}} to start each appendix. You must declare a $\backslash${\tt{section}} before using any $\backslash${\tt{subsection}} or using $\backslash${\tt{label}} ($\backslash${\tt{appendices}} by itself starts a section numbered zero.)}



\bibliographystyle{IEEEtran}
\bibliography{IEEEtran}

\end{document}